\documentclass[iicol,pdflatex,sn-vancouver,Numbered]{sn-jnl}

\usepackage{graphicx}%
\usepackage{multirow}%
\usepackage{amsmath,amssymb,amsfonts}%
\usepackage{amsthm}%
\usepackage{mathrsfs}%
\usepackage[title]{appendix}%
\usepackage{xcolor}%
\usepackage{textcomp}%
\usepackage{manyfoot}%
\usepackage{booktabs}%
\usepackage{algorithm}%
\usepackage{algorithmicx}%
\usepackage{algpseudocode}%
\usepackage{listings}%

\usepackage{booktabs}
\usepackage{graphicx}
\usepackage{amsmath}
\usepackage{amssymb}
\usepackage{booktabs}
\usepackage{multirow}
\usepackage{makecell}
\usepackage{amssymb}
\usepackage{bbm}
\usepackage{color}
\usepackage{bbding}
\usepackage{colortbl}
\usepackage{framed}

\newcommand{\keypoint}[1]{\vspace{0.1cm}\noindent\textbf{#1}}

\theoremstyle{thmstyleone}%
\theoremstyle{thmstyletwo}%

\theoremstyle{thmstylethree}%

\begin{document}

\title{A Closer Look at Conditional Prompt Tuning for Vision-Language Models}


\author[1]{\fnm{Ji} \sur{Zhang}}\email{jizhang.jim@gmail.com}

\author[2]{\fnm{Shihan} \sur{Wu}}

\author[2]{\fnm{Lianli} \sur{Gao}}

\author[3]{\fnm{Jingkuan} \sur{Song}}\email{jingkuang.song@gmail.com}

\author[4]{\fnm{Nicu} \sur{Sebe}}

\author[3]{\fnm{Heng Tao} \sur{Shen}}






\affil[1]{Southwest Jiaotong University}
\affil[2]{University of Electronic Science and Technology of China}
\affil[3]{Tongji University}
\affil[4]{University of Trento}




\abstract{
Despite the great promise of Prompt Tuning (PT) in adapting large Vision-Language Pretrained Models (VLPMs) to downstream tasks, they often struggle to overcome the Base-New Tradeoff (BNT) dilemma: as VLPMs are better tuned to a base task, their ability to generalize to new tasks diminishes.
Recent work on conditional PT addresses this problem by replacing static prompts with dynamic Visual Image Information (VII)-conditioned prompts, improving the model's generalization to new tasks to some extent.
In this work, we first identify a critical issue with existing conditional PT methods: using VII as the “condition” of prompts yields suboptimal performance, and even random noise-conditioned prompts can outperform the VII-conditioned counterparts.
On further analysis, we find that learning dynamic prompts conditioned on Textual Class Information (TCI) is the key to solving the BNT problem. 
Motivated by this, we then propose \textbf{C}lass-\textbf{a}daptive \textbf{P}rompt \textbf{T}uning (\textbf{CaPT}), which enables fast adaptation of tuned models to new classes by learning TCI-conditioned prompts from base classes.
Remarkably, CaPT can be used as a plugin to mitigate the BNT problem for existing unconditional PT schemes. Extensive experiments on 11 datasets show that CaPT consistently improves the performance of five strong unconditional PT baselines with negligible additional computational cost. 
Additionally, by integrating CaPT with our recently proposed DePT \cite{zhang2023dept} framework, we devise a new conditional PT approach, termed DeCaPT, which outperforms the H ACC of the state-of-the-art conditional PT scheme by \underline{\textbf{3.49}}\%, averaged over the 11 datasets.
Code: \url{https://github.com/Koorye/CaPT}.
}

\keywords{Prompt tuning, Conditional prompt tuning, Few-shot learning, Vision-language models.}



\maketitle

\section{Introduction}
\label{sec:introduction}
Recent years have witnessed remarkable advancements in large vision-language pretrained models (VLPMs).
Among these, the contrastive language-image pretraining (CLIP) model \cite{radford2021clip} stands out as a significant breakthrough, leveraging contrastive loss to align images and their  textual descriptions within a unified feature space. However, despite their impressive capability to capture open-set visual concepts, VLPMs exhibit a notable decline in zero-shot generalization performance when confronted with significant category, distribution, or domain shifts between upstream training data and downstream tasks \cite{luo2023closer,derakhshani2023bayesian}.

Inspired by the success of prompt engineering in natural language processing (NLP), prompt tuning (PT) has emerged as an effective and efficient paradigm to generalize VLPMs to downstream tasks \cite{zhang2023dept,khattak2023maple}. 
Given limited training data for the base (a.k.a. target) task, PT aims to optimize a task-specific prompt, which consists of a set of learnable vectors, while keeping the pretrained weights of VLPMs frozen.
Despite the great advantages, existing PT methods usually fail to escape the Base-New Tradeoff (BNT) problem. 
This means that as VLPMs are better tuned to a base task, their capability to generalize to new tasks diminishes.
Recent advance in conditional PT \cite{zhou2022conditional,derakhshani2023bayesian} addresses this problem by replacing static prompts with dynamic {Visual Image Information (VII)}-{conditioned} prompts during training, improving the generalization capability of the tuned model on new tasks to some extend.

\begin{figure}
\setlength{\abovecaptionskip}{-0.cm}  
\setlength{\belowcaptionskip}{-0.1cm} 
	\centering	
	\includegraphics[width=0.9\linewidth]{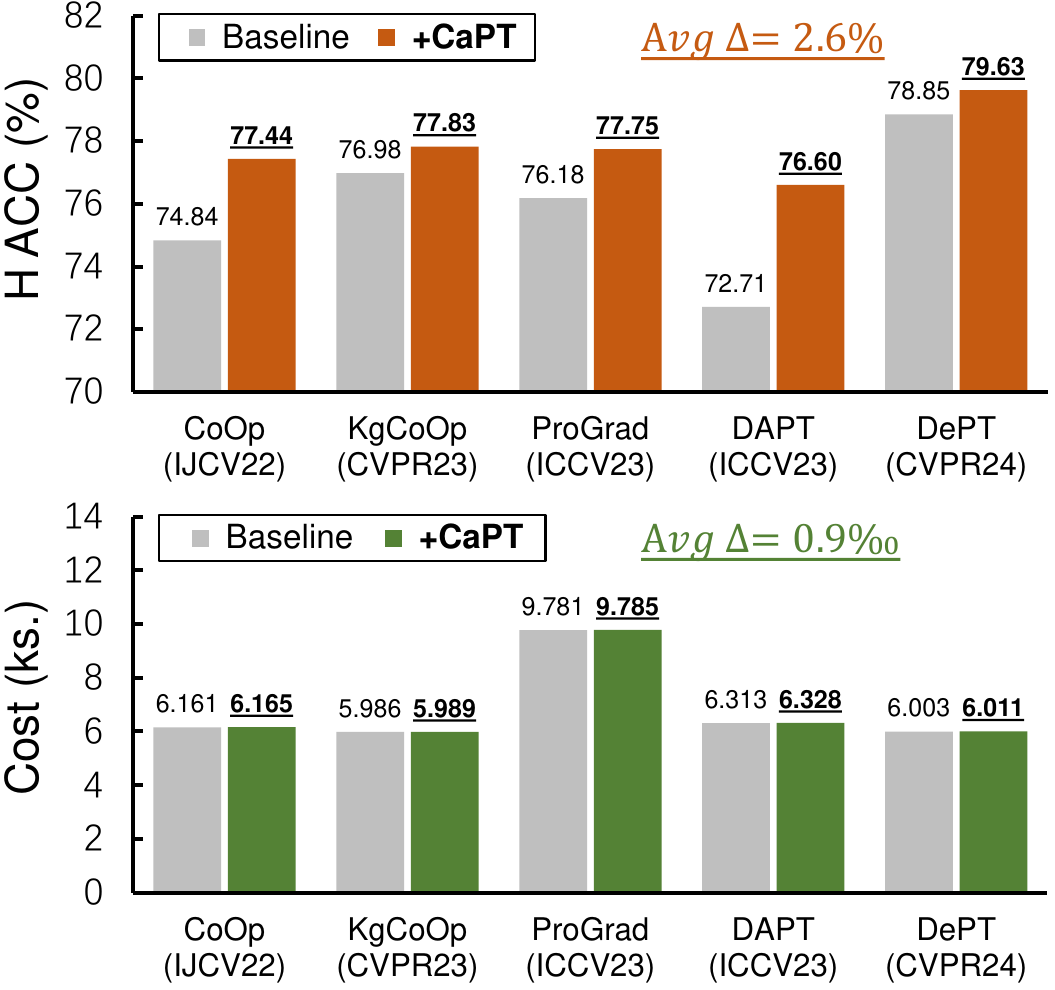} 
	\caption{\textbf{Top}: Harmonic mean of classification accuracy (H ACC) of the base and new tasks for five unconditional PT methods w/ or w/o  CaPT. The results are the average on 11 datasets in Section \ref{exp.setup}). \textbf{Down}: Computational cost of CaPT in terms of training time, with more details presented in Table \ref{cost}. As seen, CaPT can be used as a plugin to enhance existing unconditional PT approaches with negligible additional computational cost.}
	\label{Fig1}
\end{figure}
 
Over the years, it has been commonly believed that the performance enhancement of conditional PT paradigms on new tasks is primarily established by the VII injected into the prompt.  
However, our empirical results reveal that leveraging VII as the “condition” of prompts yields suboptimal performance, and even random noise-conditioned prompts can outperform the VII-conditioned counterparts (Section \ref{sect:rethink}).
On further analysis, we reveal that learning dynamic prompts conditioned on {Textual Class Information (TCI)} of intra-task classes is the key to overcoming the BNT problem of existing PT methods.
Motivated by this observation, we propose \textbf{C}lass-\textbf{a}daptive \textbf{P}rompt \textbf{T}uning (\textbf{CaPT}), which facilitates the fast adaptation of tuned models to new classes by learning TCI-conditioned prompts from base classes  (Section \ref{sect:capt}).
Specifically, CaPT uses a pretrained Word2Vec model and a learnable lightweight network (coined, Meta-Net) to capture class-adaptive textual features, which are then fused with the context vectors of the learnable prompt for conditional PT.
At inference, CaPT generates TCI-conditioned prompts for the base and new tasks by combining the extracted TCI of intra-task classes with the context vectors of the  learned prompt, respectively.
Moreover, CaPT employs a margin-adjusted Image-Text Matching (margin-ITM) loss to prevent memorization of unnecessary features of base classes during training, further enhancing the model's generalization ability on new tasks. An illustration of our proposed CaPT is presented in Fig. \ref{figpip}.

\keypoint{Flexibility and Effectiveness.} 
Our CaPT is naturally orthogonal to existing unconditional PT methods, making it flexible for improving all of them. 
We evaluate the flexibility and effectiveness of  CaPT using a broad spectrum of baselines, including CoOp \cite{zhou2022learning}, KgCoOp \cite{yao2023kgcoop}, ProGrad \cite{zhu2022prograd}, DaPT \cite{cho2023dapt} and DePT \cite{zhang2023dept}—our recently proposed state-of-the-art unconditional PT approach. 

\begin{itemize}
\item  Experimental results on 11 datasets show that CaPT consistently improves the base-to-new generalization  (Table \ref{table—B2N}), cross-dataset generalization  (Table \ref{table—CD}) and cross-domain generalization (Table \ref{table—DG}) performance of the five baselines, underscoring the strong flexibility and effectiveness of CaPT.
Notably, CaPT addresses the BNT problem for the five strong baselines in an effective and efficient manner—CaPT improves the average H ACC of the five baselines by \underline{\textbf{2.60\%}}, with only \underline{\textbf{0.9\text{\textperthousand}}} additional computational cost (Fig. \ref{Fig1}).

\item Notably, by integrating CaPT with our recently proposed DePT \cite{zhang2023dept} framework, we propose a new conditional PT scheme, named DeCaPT, which surpasses the state-of-the-art conditional PT method, i.e. Bayes \cite{derakhshani2023bayesian}, by \underline{\textbf{3.49}}\% (Table \ref{table-sota}).  
\end{itemize}

\keypoint{Contributions.} To summarize, our main contributions of this work are threefold:
\begin{itemize}

\item 
For the first time, we reveal that the key to mitigating the Base-New Tradeoff (BNT) problem in conditional prompt tuning lies in learning dynamic prompts conditioned on textual class information (TCI), rather than visual image information (VII).

\item We propose CaPT, the first TCI-conditioned PT scheme dedicated to tackling the BNT problem. Notably, CaPT can be flexibly integrated with existing unconditional PT paradigms.

\item Our extensive experiments on 11 datasets prove the flexibility and effectiveness of CaPT—CaPT consistently improves the performance of a broad spectrum of PT methods across base-to-new generalization, cross-dataset generalization and cross-domain generalization settings.

\end{itemize}

\section{Related Work}
\keypoint{Vision-Language Pretrained Models.} 
Deep learning methdos have consistently demonstrated their ability to match or even exceed human-level performance in both computer vision and natural language processing tasks \cite{jin2023context,min2023recent,khurana2023natural}. This achievement has been further amplified by the emergence of large-scale vision-language pretrained models (VLPMs). By effectively modeling the connections between image and text data, these models exhibit robust zero-shot generalization across various downstream tasks. This also underscores the profound impact of integrating multimodal information processing in advancing AI capabilities beyond traditional benchmarks.
VLPMs commonly employ three types of pretext tasks to model the semantic correspondence between vision and language modalities: {1)} image-text matching \cite{kim2021vilt,jia2021scaling}, {2)} contrastive learning \cite{huo2021wenlan,li2021align,DMG_ICML}, and {3)} masked vision/language prediction \cite{kim2021vilt,lu2019vilbert}. 
 In this study, our primary focus is on VLPMs that establish image-text alignment using contrastive learning, driven by their exceptional ability to generalize to downstream tasks. For instance, CLIP \cite{radford2021clip} achieves alignment between visual and textual features from an image encoder and a text encoder after training on 400 million text-image pairs.
Additionally, CLIP exhibits significant potential for diverse downstream applications, showcasing considerable promise for video-text retrieval \cite{ma2022x,zhu2023complementarity}, dense prediction  \cite{rao2022denseclip,zhou2022extract} and image manipulation \cite{wang2022clip,patashnik2021styleclip}. 

\keypoint{Parameter-efficient Finetuning.} 
The notable achievements of VLPMs have illuminated new possibilities, yet they also raise a pertinent question: how can we effectively transfer knowledge from VLPMs to various downstream tasks? The most straightforward approach is full-finetuning, which entails fixing the architecture of VLPMs while adjusting all parameters for the target task. Although yielding impressive results, this method becomes increasingly costly as VLPM parameter sizes expand. In response to this challenge, partial-finetuning has emerged, suggesting the updating of only a limited number of additional parameters, while maintaining the majority of pretrained parameters in a frozen state.
For instance, adapter tuning \cite{houlsby2019adapters,triantafillou2021learning,zhang2025reliable} involves the insertion of small neural modules known as adapters into each layer of the pretrained network, with only the adapters being trained during fine-tuning. Building on the achievements of prompting methods that regulate VLPMs via textual prompts, prefix tuning \cite{li2021prefix} and prompt tuning \cite{wu2025skip} introduce additional tunable prefix tokens to the input or hidden layers, focusing solely on training these soft prompts during fine-tuning for downstream tasks. Besides, the recently proposed LoRA \cite{hu2021lora} focuses on the learning of low-rank matrices to approximate parameter updates.

\keypoint{Prompt Tuning.} 
Recent research has witnessed a flourishing trend in adapting VLPMs to downstream tasks through the end-to-end learning of task-specific prompts \cite{wang2022defo,chen2022prompt,yang2025consistent,xu2025progressive,wu2024hybridprompt}. 
Given the scarcity of labeled examples during training, this approach, termed prompt tuning, aligns with the paradigm of few-shot learning \cite{zhang2023deta,fu2021meta,fu2023styleadv,zhang2022tip,zhang2023prompt}. 
 CoOp \cite{zhou2022learning} facilitates task adaptation by optimizing a set of prompt vectors within the language branch of CLIP. 
While effective, CoOp often suffers from limited generalization on new tasks due to overfitting to the base task. 
KgCoOp \cite{zhou2022conditional}  further enhances the model generalization of CoOp by minimizing the discrepancy between hand-crafted and trainable prompt tokens. ProGrad \cite{zhu2022prograd}, on the other hand, tackles overfitting by regularizing each tuning step to align with the general knowledge encoded in the hand-crafted prompt. 
In contrast to methods primarily focused on textual prompts, a growing body of work explores the integration of visual prompts for task adaptation \cite{jia2022vpt,huang2023diversity}.
By incorporating trainable prompts into both the language and text branches of CLIP, multi-model PT approaches like MaPLe\cite{khattak2023maple} and PromptSRC\cite{Khattak_2023_ICCV} demonstrate remarkable performance not only on base tasks but also on new tasks.
In our recent work \cite{zhang2023dept}, we propose Decoupled Prompt Tuning (DePT), which boosts the generalization ability of the tuned model on new tasks from a feature decoupling perspective. 
Recent work on conditional PT replaces static prompts with dynamic Visual Image Information (VII)-conditioned prompts, so as to alleviate the performance tradeoff on the base task and new tasks.
CoCoOp \cite{zhou2022conditional} introduces a lightweight meta-net to generate input-conditional tokens for each input image, thereby reducing the discrepancy between hand-crafted and trainable prompt tokens.
Bayes \cite{derakhshani2023bayesian} presents a Bayesian perspective on PT by formulating it as a variational inference problem. 
It has long been believed that the performance enhancement of conditional PT paradigms on new tasks is primarily established by the Visual Image Information (VII)-conditioned prompts.  
Nevertheless, we experimentally find that leveraging VII as the “condition” of prompts yields suboptimal performance, and even random noise-conditioned prompts can outperform the VII-conditioned counterparts.
This drives us to explore in this work what "conditions" are most conducive to model generalization in PT.

\begin{figure*}
	\centering	
\setlength{\abovecaptionskip}{-0.4cm}  
\setlength{\belowcaptionskip}{-0.2cm} 
	\includegraphics[width=1.0\linewidth]{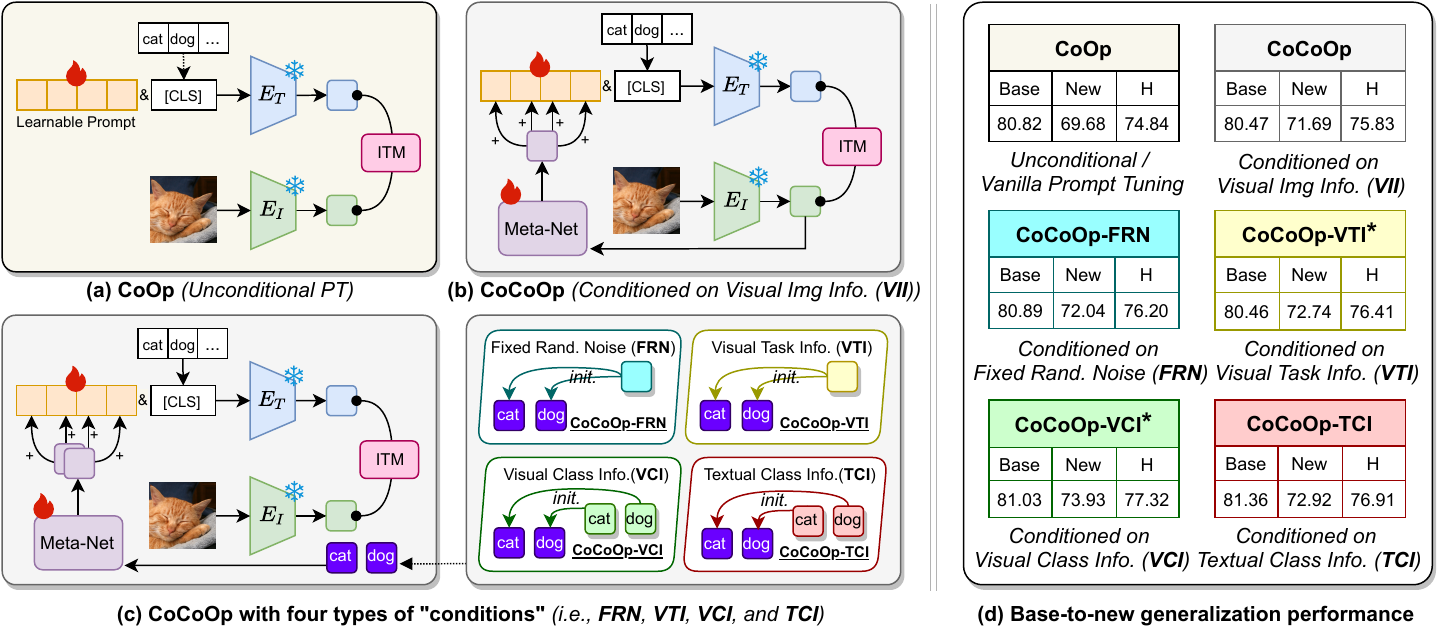} 
	\caption{\textbf{A Closer Look at Conditional Prompt Tuning}. 
         (\textbf{a}) An overview of the unconditional PT method CoOp \cite{zhou2022coop}, where $E_T$ and $E_I$ are the text encoder and image encoder of CLIP, \textit{ITM} is the image-text matching loss. (\textbf{b}) The conditional PT method CoCoOp \cite{zhou2022coop}, which uses the visual image information (\textbf{VII}) of training samples extracted by $E_I$ as the “condition” of the learnable prompt.  (\textbf{c}) CoCoOp with four types of “conditions”, including fixed random noise (\textbf{FRN}), visual task information (\textbf{VTI}), visual class information (\textbf{VCI}), and textual class information (\textbf{TCI}).  (\textbf{d}) The base-to-new generalization results of CoOp, CoCoOp and four CoCoOp variants (the results are the average on 11 datasets in Section \ref{exp.setup}). $*$ indicates CoCoOp-VTI and CoCoOp-VCI are \textit{oracle} models. \textit{Base}: base-task ACC, \textit{New}: new-task ACC. \textit{H}: the harmonic mean of \textit{Base} and \textit{New}, reflecting the performance tradeoff of the tuned model on the base task and new tasks. The larger the value of \textit{H}, the less significant the BNT problem is.}
	\label{mtv}
\end{figure*}

\section{Methodology}
\label{prelim}
In this section, we first critically analyze existing conditional prompt tuning (PT) schemes. Next, we introduce and elaborate on our proposed Class-adaptive Prompt Tuning (CaPT) approach.

\subsection{Preliminaries}
We start with an introduction of important preliminaries, including the CLIP model, the Image-Text Matching (ITM) loss, and the learning mechanism of conditional PT.

\keypoint{Contrastive Language-Image Pretraining (CLIP).}
We follow the common setup of existing approaches to perform PT on the CLIP \cite{wang2022clip} model. CLIP aims to establish an alignment between image and text features generated by an image encoder and a text encoder, respectively. 
By being exposed to 400 million image-text association pairs and employing a contrastive learning paradigm within a shared feature space, CLIP acquires a diverse array of open-set visual concepts that can be readily applied to downstream tasks.
For instance, zero-shot classification can be achieved by framing the classification task as an image-text matching problem. Initially, a prompt (e.g. “\texttt{a photo of a}”) is crafted to extract text features of all intra-task classes by presenting the class-extended prompt (e.g. “\texttt{a photo of a [CLS]}”) to the text encoder. Next, the image encoder is utilized to derive the image feature of a given input example, and class prediction is performed by comparing cosine distances between the image feature and text features of classes.

\keypoint{Prompt Tuning with an Image-Text Matching Loss.} 
The goal of PT (a.k.a. Context Optimization \cite{zhou2022coop,zhou2022conditional}) is to learn a static, task-specific prompt on a few-labeled samples from the base (or target) task. 
As shown in Fig. \ref{mtv} (\textbf{a}), a prompt is formulated as $l$ trainable context vectors, denoted as $\{\boldsymbol{v}_1,\boldsymbol{v}_2,...,\boldsymbol{v}_l\}$. 
During training, we produce the text feature of the $i$-th class by inputting the class-name-extended prompt $\boldsymbol{{c}}_i=\{\boldsymbol{v}_1,\boldsymbol{v}_2,...,\boldsymbol{v}_l,[\texttt{CLS}]\}$ to the text encoder $g(\cdot)$, where $[\texttt{CLS}]$ indicates the class name of the $i$-th class.
Denote $\boldsymbol{f}_j \in \mathbb{R}^{d}$  the image feature of a sample $\boldsymbol{x}_j$ extracted by the image encoder, the task-specific prompt can be updated via back-propagating the Image-Text Matching (ITM) loss through the frozen CLIP model. The ITM loss $\mathcal{L}_{\mathtt{ITM}}$ takes the form of
\begin{equation}
    \mathcal{L}_{\mathtt{ITM}}=-\sum_i \boldsymbol{y}_i \log \mathcal{P}(\boldsymbol{{c}}_i| \boldsymbol{x}_j),
    \label{eq.1}
\end{equation}
where $\boldsymbol{y}$ is the one-hot label, and
\begin{equation}
    \mathcal{P}(\boldsymbol{{c}}_i | \boldsymbol{x}_j)=\frac{e^{<g(\boldsymbol{{c}}_i), \boldsymbol{f}_j>/\tau}}{\sum_{t=1}^{M}e^{<g(\boldsymbol{{c}}_t), \boldsymbol{f}_j>/\tau}},
    \label{inf}
\end{equation}
 $<\cdot>$ denotes cosine similarity, $M$ is the number of classes, and $\tau$ is the temperature.

 \keypoint{Conditional Prompt Tuning.} 
Conditional PT aims to improve or at least retain the generalization capability of the tuned model (or prompt) on new tasks after learning task-specific knowledge of the base task \cite{zhou2022conditional,derakhshani2023bayesian}.
As a representative method, CoCoOp \cite{zhou2022conditional} achieves the goal by replacing the static prompt with a dynamic Visual Image Information (VII) conditioned prompt, as shown in Fig. \ref{mtv} (\textbf{b}).
Concretely, let $h_{\alpha}(\cdot)$ be a lightweight neural network parameterized by $\alpha$, coined Meta-Net, each context token in $\{\boldsymbol{v}_1,\boldsymbol{v}_2,...,\boldsymbol{v}_l\}$ can be obtained by $\boldsymbol{v}(\boldsymbol{f}_j)=\boldsymbol{v}+h_{\alpha}(\boldsymbol{f}_j)$, where $\boldsymbol{f}_j$ is the image feature of the sample $\boldsymbol{x}_j$ extracted by the image encoder of CLIP.
The image-conditioned prompt for the $i$-th class can be expressed as:
\begin{equation}
    \boldsymbol{{c}}_i(\boldsymbol{f}_j)=\{\boldsymbol{v}_1(\boldsymbol{f}_j),\boldsymbol{v}_2(\boldsymbol{f}_j),...,\boldsymbol{v}_l(\boldsymbol{f}_j),[\texttt{CLS}]\},
    \label{eq.con}
\end{equation}
and the prediction probability is calculated as
\begin{equation}
    \mathcal{P}(\boldsymbol{{c}}_i | \boldsymbol{x})=\frac{e^{(<g(\boldsymbol{{c}}_i(\boldsymbol{f}_j)), \boldsymbol{f}_j>/\tau)}}{\sum_{t=1}^{M}e^{(<g(\boldsymbol{{c}}_t(\boldsymbol{f}_j)), \boldsymbol{f}_j>/\tau)}},
    \label{inf22}
\end{equation}
During conditional prompt tuning, the context vectors  $\{\boldsymbol{v}_1,\boldsymbol{v}_2,...,\boldsymbol{v}_l\}$ together with the parameters of the Meta-Net, i.e. $\alpha$, are updated.
At inference, CoCoOp makes the prompt conditioned on the VII of each input instance (image) of the base/new task rather than fixed once learned. 

\subsection{A Closer Look at Conditional Prompt Tuning}
\label{sect:rethink}
Over the past years, it has long been believed that the remarkable performance enhancement of conditional PT paradigms on new tasks is
primarily established by the extracted visual image information (VII) from input instances \cite{zhou2022conditional,derakhshani2023bayesian}.
In this section, we present an insightful analysis of the effectiveness of VII and other “conditions” for conditional PT. 
To this end, we use CoCoOp \cite{zhou2022conditional} as the baseline and replace its VII-conditioned prompt with a $\mathcal{X}$-conditioned counterpart. 
Specifically, we consider the following variants of $\mathcal{X}$:

\keypoint{{i) Fixed Random Noise (FRN)}.} 
During training on the base task, the learnable prompt is conditioned on fixed random noise (FRN), denoted as $\boldsymbol{\zeta} \in \mathbb{R}^{d}$, for \textit{all} training samples. 
Hence, we have
\begin{equation}
    \boldsymbol{{c}}_i({\boldsymbol{\zeta}})=\{\boldsymbol{v}_1(\boldsymbol{\zeta}),\boldsymbol{v}_2(\boldsymbol{\zeta}),...,\boldsymbol{v}_l(\boldsymbol{\zeta}),[\texttt{CLS}]\},
\end{equation}
where $\boldsymbol{v}(\boldsymbol{\zeta})=\boldsymbol{v}+h_{\alpha}(\boldsymbol{\zeta})$. 
At inference, the learned Meta-Net takes the same $\boldsymbol{\zeta}$ input to construct FRN-conditioned prompts for both base and new tasks.
In a nutshell, the VII of all intra-task samples for CoCoOp are replaced with the same FRN (i.e. $\boldsymbol{\zeta}$) for the base task and new tasks. We denote this variant of CoCoOp as CoCoOp-FRN. 

\keypoint{{ii) Visual Task Information (VTI)}.} 
The VII for CoCoOp is represented by instance-level image features extracted from the CLIP's image encoder. In contrast, Visual Task Information (VTI) refers to the task prototype extracted by the image encoder.
Let $\{\boldsymbol{f}_{j}\}_{j=1}^N$ be a set of visual features of training samples from the base task, we compute the VTI of the base task as the average of all intra-task image features:
\begin{equation}
    \boldsymbol{\mu}_{base} = \frac{1}{N} \sum_{{j}=1}^N \boldsymbol{f}_{j}.
\end{equation}
During training, the learnable prompt is conditioned on $\boldsymbol{\mu}_{b}$ for all training samples $\boldsymbol{x}_j, j=1,...,N$. Therefore, we have
\begin{equation}
    \boldsymbol{{c}}_i(\boldsymbol{\mu}_{base})=\{\boldsymbol{v}_1(\boldsymbol{\mu}_{base}),...,\boldsymbol{v}_l(\boldsymbol{\mu}_{base}),[\texttt{CLS}]\},
\end{equation}
where $\boldsymbol{v}(\boldsymbol{\mu}_{base})=\boldsymbol{v}+h_{\alpha}(\boldsymbol{\mu}_{base})$. 
For each new task, denote $\boldsymbol{\mu}_{new}$ and $\boldsymbol{{c}}_i(\boldsymbol{\mu}_{new})$ similar to $\boldsymbol{\mu}_{base}$ and $\boldsymbol{{c}}_i(\boldsymbol{\mu}_{base})$. 
At inference, the learned Meta-Net takes $\boldsymbol{\mu}_{base}$ (resp. $\boldsymbol{\mu}_{new}$)  as input to construct a VTI-conditioned prompt and build a task-specific classifier for the base (resp. new) task. 
Here, we denote this variant of CoCoOp as CoCoOp-VTI\footnote{Since a set of training samples for each new task is not available during PT on the base task, CoCoOp-VTI, as well as the subsequently mentioned CoCoOp-VCI, are considered \textit{oracle} models.}.

\keypoint{{iii) Visual Class Information (VCI)}.}  
Visual Class Information (VCI) is defined as the intra-task class prototypes produced by the CLIP's image encoder.
Let $C_i = \{\boldsymbol{f}_{j}\}_{j=1}^{N_{i}}$ be the extracted image features of training samples of the $i$-th base class, we compute the VCI for class $i$, $\boldsymbol{p}_{base}^i$, by
\begin{equation}
    \boldsymbol{p}_{base}^i = \frac{1}{N_i} \sum_{\boldsymbol{f}_{j} \in C_i} \boldsymbol{f}_{j}, \quad i=1,...,M_{base},
\end{equation}
where $M_{base}$ is the number of base classes, and $N_i$ is the number of training samples in class $i$.
During training on the base task, for each sample $\boldsymbol{x}_j$ from class $i$, the learnable prompt is conditioned on the corresponding VCI: $\boldsymbol{\mu}_{base}^i$. Thus, we have
\begin{equation}
    \boldsymbol{{c}}_i(\boldsymbol{p}_{base}^i)=\{\boldsymbol{v}_1(\boldsymbol{p}_{base}^i),...,\boldsymbol{v}_l(\boldsymbol{p}_{base}^i),[\texttt{CLS}]\},
    \label{eq.base}
\end{equation}
where $\boldsymbol{v}_l(\boldsymbol{\mu}_{base})=\boldsymbol{v}_l+h_{\alpha}(\boldsymbol{\mu}_{base})$. 
For each class $i$ from the new task, denote $\boldsymbol{p}_{new}^i$ and $\boldsymbol{{c}}_i(\boldsymbol{p}_{new}^i)$ similar to $\boldsymbol{p}_{base}^i$ and $\boldsymbol{{c}}_i(\boldsymbol{p}_{base}^i)$.
At inference, Meta-Net takes $\boldsymbol{p}_{base}^i$ (resp. $\boldsymbol{p}_{new}^i$ ) as input to construct a VCI-conditioned prompt for the $i$-th base (resp. new) class—leveraging
Eq. \ref{eq.base} for the base task and $\boldsymbol{{c}}_i(\boldsymbol{p}_{new}^i)=\{\boldsymbol{v}_1(\boldsymbol{p}_{new}^i),...,\boldsymbol{v}_l(\boldsymbol{p}_{new}^i),[\texttt{CLS}]\}$ for the new task.
We denote this variant of CoCoOp as CoCoOp-VCI.

\keypoint{{iv) Textual Class Information (TCI)}.} 
When performing PT on the base task, we do not have training samples available to calculate either the VCI or the VTI for new tasks. In other words, the aforementioned CoCoOp-VTI and CoCoOp-VCI are thus considered \textit{oracle} models.
Fortunately, we have access to the class names of both the base task and new tasks, which can be readily employed to generate  Textual Class Information (TCI) of intra-task classes using a pretrained textual feature extractor. 
Let $\{\texttt{C}_{base}^{i}\}_{i=1}^{{M_{base}}}$ be the set of class names of the base task, we denote the TCI of the $i$-th class in the base task as follows:
\begin{equation}
    \boldsymbol{e}_{base}^i = \varphi(\texttt{C}_{base}^{i}), \quad i=1,...,M_{base},
\end{equation}
where $\varphi(\cdot)$ is a textual feature extractor, e.g., pretrained Tokenizers \cite{joulin2016fasttext,yamada2020wikipedia2vec,pennington2014glove} or text encoder of CLIP \cite{gao2024clip}.
During training on the base task, for each sample $\boldsymbol{x}_j$ from class $i$, the learnable prompt is conditioned on $\boldsymbol{e}_{base}^i$:
\begin{equation}
    \boldsymbol{{c}}_i(\boldsymbol{e}_{base}^i)=\{\boldsymbol{v}_1(\boldsymbol{e}_{base}^i),...,\boldsymbol{v}_l(\boldsymbol{e}_{base}^i),[\texttt{C}_{base}^{i}]\},
\end{equation}
where $\boldsymbol{v}_l(\boldsymbol{\mu}_{baase})=\boldsymbol{v}_l+h_{\alpha}(\boldsymbol{\mu}_{base})$. 
For each class $i$ from the new task, denote $\boldsymbol{e}_{new}^i$ and $\boldsymbol{{c}}_i(\boldsymbol{e}_{new}^i)$ similar to $\boldsymbol{e}_{base}^i$ and $\boldsymbol{{c}}_i(\boldsymbol{e}_{base}^i)$.
At inference, the learned Meta-Net takes $\boldsymbol{e}_{base}^i$ (resp. $\boldsymbol{e}_{new}^i$ ) as input to construct a TCI-conditioned prompt for the $i$-th base (resp. new) class, using
Eq. \ref{eq.base} for the base task and $\boldsymbol{{c}}_i(\boldsymbol{e}_{new}^i)=\{\boldsymbol{v}_1(\boldsymbol{e}_{new}^i),...,\boldsymbol{v}_l(\boldsymbol{e}_{new}^i),[\texttt{CLS}]\}$ for the new task.
In this section, $\varphi(\cdot)$ specifically refers to the pretrained textual encoder of CLIP. We denote this variant of CoCoOp as CoCoOp-TCI.

\keypoint{Analysis.} 
It is interesting to scrutinize the effectiveness of different “conditions” for conditional PT. 
To this end, we regard CoCoOp \cite{zhou2022conditional} as the baseline and replace its VII-conditioned prompt with FRN-, VTI-, VCI-, and TCI-conditioned prompts, respectively.
The obtained base-to-new generalization performance of CoOp, CoCoOp, as well as the four variants of CoCoOp (i.e., CoCoOp-FRN, CoCoOp-VTI, CoCoOp-VCI and CoCoOp-TCI) are reported in Fig. \ref{mtv} (\textbf{d}), where the results are the average over 11 datasets in Section \ref{exp.setup}. From the obtained results, we have the following important observations.

\begin{itemize}
\item [1)] \textbf{CoCoOp vs. CoCoOp-FRN}. Surprisingly, leveraging fixed random noise (FRN) as “condition”, CoCoOp-FRN achieves better generalization results than CoCoOp on both the base task and new tasks, which reveals that using the visual image information (VII) as the “condition” of learnable prompts yields suboptimal base-to-new generalization performance.

\item [2)]  \textbf{CoCoOp vs. CoCoOp-VTI and CoCoOp-VCI}. The two \textit{oracle} models CoCoOp-VTI and CoCoOp-VCI consistently outperform CoCoOp by significant  margins, which demonstrates the superior value of visual task information (VTI) and visual class information (VCI) over VII for tackling the BNT problem. 

\item [3)]  \textbf{CoCoOp-VCI vs. CoCoOp-VTI}. CoCoOp-VCI outperforms CoCoOp-VTI and CoCoOp, revealing that incorporating class-level information to the prompt is of greater importance for conditional PT.

\item [4)]  \textbf{CoCoOp-TCI vs. CoCoOp-VCI}. By replacing \textit{visual} class information (VCI) in CoCoOp-VCI with \textit{textual} class information (TCI), CoCoOp-TCI achieves performance close to the \textit{oracle} results established by CoCoOp-VCI on both the base task and new tasks.

\end{itemize}

\begin{figure}
\setlength{\abovecaptionskip}{0.15cm}  
\setlength{\belowcaptionskip}{-0.2cm} 
	\centering	
	\includegraphics[width=0.8\linewidth]{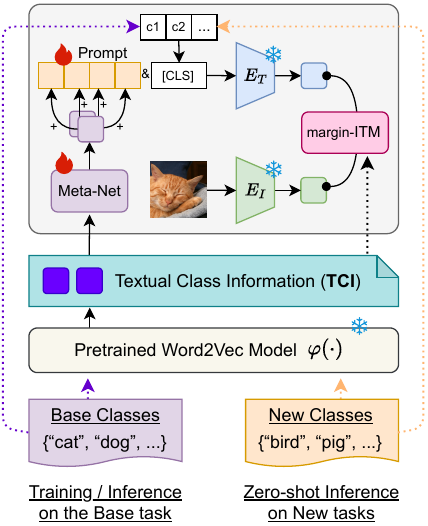} 
	\caption{Overview of the devised Class-adaptive Prompt Tuning (\textbf{{CaPT}}). 
         CaPT uses a pretrained Word2Vec model to produce the textual class information (\textbf{TCI}) of intra-task classes, which are then fed into a Meta-Net to obtain the “condition” of the prompt for conditional PT. 
         During training on the base task, a {margin-ITM} loss is also devised based on the obtained TCI of base classes, aiming to boost the generalizability of the tuned model on new tasks by avoiding memorization of unnecessary base class features.}
	\label{figpip}
\end{figure}

\subsection{CaPT: Class-adaptive Prompt Tuning}
\label{sect:capt}
Section \ref{sect:rethink} demonstrates that learning dynamic prompts conditioned on textual class information (TCI)—rather than visual image information (VII), is the key to mitigating the BNT problem.
Motivated by this, in this work we propose \textbf{C}lass-\textbf{a}daptive \textbf{P}rompt \textbf{T}uning (\textbf{CaPT}), which enables fast adaptation of tuned models to new classes by learning TCI-conditioned prompts from base classes, as shown in Fig. \ref{figpip}. 

Specifically, CaPT enables fast adaptation of the tuned model (or prompt) on new tasks/classes by learning TCI-conditioned prompts on the base task. 
Let $\{\texttt{C}_{base}^{i}\}_{i=1}^{{M_{base}}}$ be the set of class names of the base task, we first compute the TCI of the $i$-th base class as follows:
\begin{equation}
    \boldsymbol{e}_{base}^i = \varphi(\texttt{C}_{base}^{i}), \quad i=1,...,M_{base},
\end{equation}
where $\varphi(\cdot)$ is the pretrained Word2Vec model, which transforms class names to word tokens; $M_{base}$ is the number of base classes.
During training on the base task, for each sample $\boldsymbol{x}_j$ from class $i$, the learnable prompt is conditioned on $\boldsymbol{e}_{base}^i$:
\begin{equation}
    \boldsymbol{{c}}_i(\boldsymbol{e}_{base}^i)=\{\boldsymbol{v}_1(\boldsymbol{e}_{base}^i),...,\boldsymbol{v}_l(\boldsymbol{e}_{base}^i),[\texttt{C}_{base}^{i}]\},
\end{equation}
where $\boldsymbol{v}_l(\boldsymbol{e}_{base}^i)=\boldsymbol{v}_l+h_{\alpha}(\boldsymbol{e}_{base}^i)$, and $h_{\alpha}(\cdot)$ is a lightweight Meta-Net (e.g., MLP) parameterized by $\alpha$. 
Denote $\boldsymbol{f}_j \in \mathbb{R}^{d}$ as the image feature of $\boldsymbol{x}_j$ extracted by the image encoder, the prediction probability is calculated as
\begin{equation}
    \mathrm{p}_0(\boldsymbol{{c}}_i | \boldsymbol{x}_j)=\frac{e^{(<g(\boldsymbol{{c}}_i(\boldsymbol{e}_{base}^i)), \boldsymbol{f}_j>/\tau)}}{\sum_{t=1}^{M}e^{(<g(\boldsymbol{{c}}_t(\boldsymbol{e}_{base}^t)), \boldsymbol{f}_j>/\tau)}}.
    \label{inf2}
\end{equation}
During training, the pretrained parameters of the tokenizer (or text encoder) and CLIP are frozen, and only the parameters of the prompt (i.e. the context tokens: $\{\boldsymbol{v}_1,\boldsymbol{v}_2,...,\boldsymbol{v}_l\}$) and the Meta-Net (i.e. $\alpha$) are updated—using the ITM loss $\mathcal{L}_{\mathtt{ITM}}$ in Eq. \ref{eq.1}. 
Denote $\{\boldsymbol{v}_1^*,\boldsymbol{v}_2^*,...,\boldsymbol{v}_l^*\}$ as the updated context tokens of the prompt and $\alpha^*$ as the learned parameters of the Meta-Net.

During inference on the base task, we calculate the TCI-conditioned prompt for the $i$-th class in the base task as:
\begin{equation}
    \boldsymbol{{c}}_i(\boldsymbol{e}_{base}^i)=\{\boldsymbol{v}_1^*(\boldsymbol{e}_{base}^i),...,\boldsymbol{v}_l^*(\boldsymbol{e}_{base}^i),[\texttt{C}_{base}^{i}]\},
\end{equation}
where $\boldsymbol{v}_l^*(\boldsymbol{e}_{base}^i)=\boldsymbol{v}_l^*+h_{\alpha^*}(\boldsymbol{e}_{base}^i)$.
Similarly, when performing zero-shot inference on new tasks, we can compute the TCI-conditioned prompt for the $k$-th new class as:
\begin{equation}
    \boldsymbol{{c}}_k(\boldsymbol{e}_{new}^k)=\{\boldsymbol{v}_1^*(\boldsymbol{e}_{new}^k),...,\boldsymbol{v}_l^*(\boldsymbol{e}_{new}^k),[\texttt{C}_{new}^{k}]\},
\end{equation}
where $\boldsymbol{v}_l^*(\boldsymbol{e}_{new}^k)=\boldsymbol{v}_l^*+h_{\alpha^*}(\boldsymbol{e}_{new}^k)$, $\texttt{C}_{new}^{k}$ is the class name of the $k$-th new class,  $\boldsymbol{e}_{new}^k$ is the TCI of the $k$-th new class, and $k=1,...,M_{new}$.
In this way, we can leverage the obtained $\{\boldsymbol{{c}}_i(\boldsymbol{e}_{base}^i)\}_{i=1}^{M_{base}}$ and  $\{\boldsymbol{{c}}_k(\boldsymbol{e}_{new}^k)\}_{k=1}^{M_{new}}$ to  
construct task-specific classifiers for predicting the categories of testing images from the base task and new tasks, respectively.

\begin{figure}[t]
\centering	
\setlength{\abovecaptionskip}{0.1cm}  
\setlength{\belowcaptionskip}{-0.2cm} 
	\includegraphics[width=0.98\linewidth]{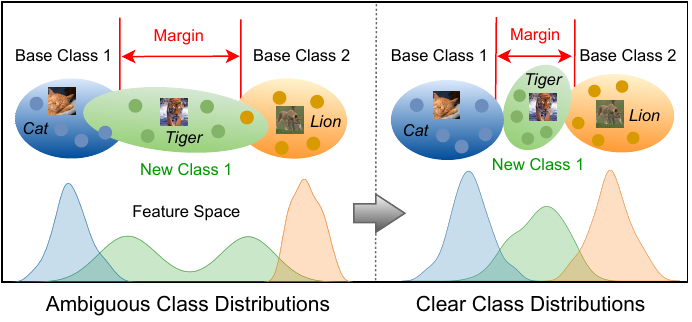} 
	\caption{Motivation of our devised Margin-adjusted Image-Text Matching (\textbf{margin-ITM}) loss in CaPT: excessively enlarging the margin between similar base classes damages the capacity of the tuned model to represent new classes.}
	\label{figmitm}
\end{figure}

\keypoint{Margin-adjusted Image-Text Matching.} 
To further enhance the generalization ability of the tuned model on new tasks, we devise a novel Margin-adjusted Image-Text Matching (margin-ITM) loss for PT. The motivation behind margin-ITM is that excessively enlarging the margin between similar base classes during PT can impair the model's generalizability to represent new classes, as illustrated in Fig. \ref{figmitm}.
Furthermore, the obtained TCI of base classes can be systematically explored to model the inter-class relationships and guide the PT process.

Specifically, given the TCI of base classes $\{\boldsymbol{e}_{base}^i\}_{i=1}^{M_{base}}$, the similarity of each pair of base classes can be computed by $s_{uv}=\epsilon\times<\boldsymbol{e}_{base}^u,\boldsymbol{e}_{base}^v>$, where $<\cdot>$ denotes cosine similarity, $\epsilon$ is a hyperparameter.
According to Eq. \ref{inf2}, the prediction probability of a training sample $\boldsymbol{x}_j$ belonging to the $i$-th base class, $\mathcal{P}^{\prime}(\boldsymbol{{c}}_i | \boldsymbol{x}_j)$, can be calculated as:
\begin{equation}
    \mathrm{p}(\boldsymbol{{c}}_i | \boldsymbol{x}_j) = 
    \frac{e^{(<g(\boldsymbol{{c}}_i(\boldsymbol{e}_{base}^i)), \boldsymbol{f}_j>/\tau)}}{\sum_{i^{\prime}=1}^{M_{base}}e^{(<g(\boldsymbol{{c}}_{i^{\prime}}(\boldsymbol{e}_{base}^{i^{\prime}})), \boldsymbol{f}_j>-s_{i^{\prime}i}/\tau)}}.
\label{infmargin}
\end{equation}
Thus, the margin-ITM loss can be expressed as
\begin{equation}
    \mathcal{L}_{\mathtt{margin-ITM}}=-\sum_i \boldsymbol{y}_i \log \mathrm{p}(\boldsymbol{{c}}_i | \boldsymbol{x}_j),
    \label{eq.margin-itm}
\end{equation}
where $\boldsymbol{y}$ is the one-hot label. 
By exploiting the TCI of base classes properly, our margin-ITM loss avoids the samples from similar classes to be excessively separable in the embedding space. The more generalizable embedding space will help
better recognize samples from new classes.

\section{Experiments}
In this section, we first validate the flexibility and effectiveness of CaPT framework by applying it to five unconditional PT schemes. 
Then, we present ablation studies to analyze the impacts of different factors on CaPT.
At the end, we investigate the computational efficiency of CaPT.

\subsection{Experimental Setup}
\label{exp.setup}
\keypoint{Baselines.}
We apply our CaPT to a broad spectrum of baseline approaches, including CoOp \cite{zhou2022learning}, CoCoOp \cite{zhou2022conditional}, KgCoOp \cite{yao2023kgcoop}, ProGrad \cite{zhu2022prograd}, DAPT \cite{cho2023dapt} and DePT \cite{zhang2023dept}. In particular, our recently proposed Decoupled Prompt Tuning (DePT) framework addressed the BNT problem from a feature decoupling perspective, and established state-of-the-art performance on several PT benchmarks. For fair comparisons, all those baselines with or without our CaPT are trained (or reproduced) for 10 epochs in our experiments. 

\keypoint{Datasets}. 
We conduct experiments on datasets from diverse sources. Concretely, for the settings of {{base-to-new generalization}} and {cross-dataset generalization}, we employ {11} datasets: ImgNet \cite{deng2009imagenet}, Caltech \cite{fei2004learning}, OxfordPets \cite{parkhi2012cats}, StanfordCars \cite{krause20133d}, Flowers \cite{nilsback2008automated},
Food101 \cite{bossard2014food}, FGVCAircraft \cite{maji2013fine}, EuroSAT \cite{helber2019eurosat}, UCF101 \cite{soomro2012ucf101}, DTD \cite{cimpoi2014describing}, and SUN397 \cite{xiao2010sun};
for the {domain generalization} setting, we utilize ImgNet as the source domain (i.e. the base task), and its four variants ImgNet-V2 \cite{recht2019imagenet}, ImgNet-Sketch \cite{wang2019learning}, ImgNet-A \cite{gao2022generating} and ImgNet-R \cite{hendrycks2021many} as target domains (i.e. new tasks).

\keypoint{Evaluation Metric.}
We report base-task accuracy (\textit{Base}), new-task accuracy (\textit{New}) and their harmonic mean (\textit{H}) to compare the performance of different methods.
The larger the value of \textit{H}, the less significant the BNT problem is \cite{zhou2022conditional,zhu2022prograd,khattak2023promptsrc}.
All results the average of 3 runs with different seeds. 

\keypoint{Implementation Details.}
Our implementation of CaPT is based on the open-source Github repository of DePT \cite{zhang2023dept}\footnote{\url{https://github.com/Koorye/DePT}}.
For fair comparisons, when applying CaPT to those baseline methods, we use the same experimental setup (e.g., feature backbone, prompt length and learning rate) as used in their original implementations. 
All baselines with or without CaPT are consistently trained (or reproduced) for 10 epochs in our experiments. 
We use the pretrained text encoder of CLIP as the Word2Vec model $\varphi(\cdot)$.
The structure of Meta-Net is a two-layer MLP, and the bottleneck dimension is set to 32.
We adjust the scaling weight $\epsilon$ (in the margin-ITM head) and the number of training epochs for our CaPT in ablation studies. 
The above hyperparameters are fixed across all datasets.
Without otherwise specified, every base task is formulated as a many-way 16-shot task.
All experimental results are the average over 3 runs with different seeds.
We conduct experiments using a {NVIDIA V100} GPU. 

\begin{table*}[htbp]
\setlength{\abovecaptionskip}{0.cm}  
\setlength{\belowcaptionskip}{0.1cm} 
  \centering 
  \tabcolsep 0.056in
  \footnotesize
  \caption{Base-to-new generalization performance of state-of-the-art PT methods w/ or w/o our CaPT on 11 datasets. }
    \begin{tabular}{l|ccc|ccc|ccc|ccc}
       \hline
     \multicolumn{1}{l|}{\multirow{2}{*}{\makecell[c]{{Method}}}}   & \multicolumn{3}{c|}{\cellcolor{green!8}{\textbf{Avg on 11 Datasets}}} & \multicolumn{3}{c|}{\cellcolor{gray!0} {{{ImageNet}}}}  & \multicolumn{3}{c|}{\cellcolor{gray!0}{Caltech101}} & \multicolumn{3}{c}{\cellcolor{gray!0}{OxfordPets}}  \\
    \cline{2-13} 
    & \multicolumn{1}{c}{Base} & \multicolumn{1}{c}{New} & \multicolumn{1}{c|}{H} & \multicolumn{1}{c}{Base} & \multicolumn{1}{c}{New} & \multicolumn{1}{c|}{H} &  \multicolumn{1}{c}{Base} & \multicolumn{1}{c}{New} & \multicolumn{1}{c|}{H}  & \multicolumn{1}{c}{Base} & \multicolumn{1}{c}{New} & \multicolumn{1}{c}{H} \\
     \hline
     CoOp  \cite{zhou2022coop}, \textit{IJCV22} &80.82&69.68&74.84 &76.60&\textbf{69.70}&\textbf{72.99}&\textbf{98.03}&93.83&95.89&\textbf{95.53}&97.40&\textbf{96.46}\\
    \textbf{+CaPT}(\textit{Ours}) &\textbf{81.32}&\textbf{73.92}&\textbf{77.44}&\textbf{76.73}&69.60&\textbf{72.99}&97.97&\textbf{94.67}&\textbf{96.29}&94.83&\textbf{97.47}&96.13\\
    \hline
         {KgCoOp} \cite{yao2023kgcoop}, \textit{CVPR23} &81.08&73.28&76.98&76.27&\textbf{70.47}&\textbf{73.25}&\textbf{97.80}&94.27&96.00&\textbf{95.30}&\textbf{97.80}&\textbf{96.53}\\
     \textbf{+CaPT}(\textit{Ours})  &\textbf{81.50}&\textbf{74.47}&\textbf{77.83}&\textbf{76.60}&69.63&72.95&97.93&\textbf{94.40}&\textbf{96.13}&94.93&97.50&96.20\\
    \hline
         ProGrad\cite{zhu2022prograd}, \textit{ICCV23} &79.33&73.27&76.18&74.73&\textbf{69.70}&72.13&98.03&\textbf{95.47}&\textbf{96.73}&94.87&97.67&96.25\\
    \textbf{+CaPT}(\textit{Ours})   &\textbf{80.12}&\textbf{75.51}&\textbf{77.75}&\textbf{75.63}&\textbf{69.70}&\textbf{72.55}&\textbf{98.13}&94.93&96.51&\textbf{95.03}&\textbf{97.83}&\textbf{96.41}\\
    \hline
         {DAPT} \cite{cho2023dapt}, \textit{ICCV23}  &80.58&66.23&72.71&76.77&\textbf{69.70}&73.06&\textbf{97.90}&\textbf{94.37}&\textbf{96.10}&95.57&\textbf{97.47}&96.51\\
     \textbf{+CaPT}(\textit{Ours}) &\textbf{80.75}&\textbf{72.86}&\textbf{76.60}&\textbf{76.63}&69.63&\textbf{72.97}&97.80&93.93&95.83&\textbf{95.83}&97.40&\textbf{96.61}\\
    \hline
         {DePT} \cite{zhang2023dept}, \textit{CVPR24}(\textit{Ours})  &83.64&74.58&78.85&\textbf{77.13}&\textbf{70.27}&\textbf{73.54}&98.30&\textbf{94.67}&96.45&\textbf{94.37}&97.23&\textbf{95.78}\\
     \textbf{+CaPT}(\textit{Ours}) &\textbf{83.99}&\textbf{75.70}&\textbf{79.63}&77.00&70.00&73.33&\textbf{98.40}&94.60&\textbf{96.46}&93.87&\textbf{97.57}&95.68\\
     \hline
     \multicolumn{1}{l|}{\multirow{2}{*}{\makecell[c]{{Method}}}}  & \multicolumn{3}{c|}{\cellcolor{gray!0}{StanfodCars}}  & \multicolumn{3}{c|}{\cellcolor{gray!0}{Flowers102}}  & \multicolumn{3}{c|}{\cellcolor{gray!0}{Food101}} & \multicolumn{3}{c}{\cellcolor{gray!0}{FGVCAircraft}}  \\
    \cline{2-13} 
    &  \multicolumn{1}{c}{Base} & \multicolumn{1}{c}{New} & \multicolumn{1}{c|}{H} &  \multicolumn{1}{c}{Base} & \multicolumn{1}{c}{New} & \multicolumn{1}{c|}{H}  &  \multicolumn{1}{c}{Base} & \multicolumn{1}{c}{New} & \multicolumn{1}{c|}{H}  &  \multicolumn{1}{c}{Base} & \multicolumn{1}{c}{New} & \multicolumn{1}{c}{H}  \\
     \hline
     CoOp  \cite{zhou2022coop}, \textit{IJCV22}  &72.00&68.13&70.01&\textbf{96.33}&72.30&82.60&\textbf{90.50}&\textbf{91.77}&\textbf{91.13}&34.13&16.80&22.52\\
    \textbf{+CaPT}(\textit{Ours})   &\textbf{74.07}&\textbf{74.07}&\textbf{74.07}&96.83&\textbf{74.57}&\textbf{84.25}&90.43&91.43&90.93&\textbf{35.37}&\textbf{34.17}&\textbf{34.76}\\
    \hline
         {KgCoOp} \cite{yao2023kgcoop}, \textit{CVPR23} &71.53&\textbf{75.10}&73.27&95.37&\textbf{74.53}&83.67&\textbf{90.60}&\textbf{91.63}&\textbf{91.11}&35.60&\textbf{34.97}&35.28\\
     \textbf{+CaPT}(\textit{Ours})  &\textbf{74.40}&74.17&\textbf{74.28}&\textbf{97.40}&\textbf{74.53}&\textbf{84.45}&90.50&91.57&91.03&\textbf{36.63}&34.83&\textbf{35.71}\\
    \hline
         ProGrad \cite{zhu2022prograd}, \textit{ICCV23}  &69.83&\textbf{75.13}&72.39&92.13&74.50&82.38&\textbf{90.93}&\textbf{91.97}&\textbf{91.45}&33.97&24.73&28.62\\
     \textbf{+CaPT}(\textit{Ours})   &\textbf{71.57}&75.07&\textbf{73.27}&\textbf{94.93}&\textbf{75.20}&\textbf{83.92}&90.80&91.80&91.30&\textbf{35.03}&\textbf{35.90}&\textbf{35.46}\\
    \hline
         {DAPT} \cite{zhang2023dept}, \textit{ICCV23} &72.37&74.50&73.42  &\textbf{96.37}&71.53&82.11&90.23&91.17&90.70&28.50&23.13&25.65\\
     \textbf{+CaPT}(\textit{Ours}) &\textbf{72.90}&\textbf{74.00}&\textbf{73.45}&96.30&\textbf{73.10}&\textbf{83.11}&\textbf{90.43}&\textbf{91.30}&\textbf{90.86}&\textbf{36.70}&\textbf{30.60}&\textbf{33.37}\\
    \hline
         {DePT}, \textit{CVPR24}(\textit{Ours}) &79.23&74.57&76.83&97.97&\textbf{75.50}&85.28&\textbf{90.50}&\textbf{91.70}&\textbf{91.10}&42.90&35.03&38.57\\
     \textbf{+CaPT}(\textit{Ours}) &\textbf{79.87}&\textbf{74.77}&\textbf{77.23}&\textbf{98.53}&75.43&\textbf{85.45}&90.47&91.63&91.05&\textbf{43.47}&\textbf{35.93}&\textbf{39.34}\\
    \hline
     \multicolumn{1}{l|}{\multirow{2}{*}{\makecell[c]{{Method}}}} & \multicolumn{3}{c|}{\cellcolor{gray!0}{SUN397}}    & \multicolumn{3}{c|}{\cellcolor{gray!0}{DTD}}  & \multicolumn{3}{c|}{\cellcolor{gray!0}{EuroSAT}} & \multicolumn{3}{c}{\cellcolor{gray!0}{UCF101}}   \\
    \cline{2-13} 
    &  \multicolumn{1}{c}{Base} & \multicolumn{1}{c}{New} & \multicolumn{1}{c|}{H}& \multicolumn{1}{c}{Base} & \multicolumn{1}{c}{New} & \multicolumn{1}{c|}{H} &  \multicolumn{1}{c}{Base} & \multicolumn{1}{c}{New} & \multicolumn{1}{c|}{H}  &  \multicolumn{1}{c}{Base} & \multicolumn{1}{c}{New} & \multicolumn{1}{c}{H}  \\
     \hline
     CoOp  \cite{zhou2022coop}, \textit{IJCV22} &80.40&75.90&78.09 &\textbf{78.70}&49.47&60.75&83.73&58.70&69.02&83.10&72.43&77.40\\
     \textbf{+CaPT}(\textit{Ours})   &\textbf{81.57}&\textbf{76.90}&\textbf{79.16}&77.87&\textbf{55.40}&\textbf{64.74}&\textbf{85.03}&\textbf{69.00}&\textbf{76.18}&\textbf{83.87}&\textbf{75.83}&\textbf{79.65}\\
    \hline
         {KgCoOp} \cite{yao2023kgcoop}, \textit{CVPR23} &80.93&76.47&78.64&\textbf{79.33}&\textbf{58.03}&\textbf{67.03}&\textbf{85.43}&57.17&68.50&\textbf{83.70}&75.67&79.48\\
     \textbf{+CaPT}(\textit{Ours}) &\textbf{81.93}&\textbf{77.40}&\textbf{79.60}  &79.23&54.73&64.74&83.60&\textbf{72.33}&\textbf{77.56}&83.33&\textbf{78.07}&\textbf{80.61}\\
    \hline
         ProGrad \cite{zhu2022prograd}, \textit{ICCV23}  &78.10&77.87&77.98&74.90&57.13&64.82&\textbf{84.20}&67.03&74.64&80.97&74.73&77.73\\
    \textbf{+CaPT}(\textit{Ours})   &\textbf{79.77}&\textbf{77.90}&\textbf{78.82}&\textbf{75.60}&\textbf{58.37}&\textbf{65.87}&82.77&\textbf{75.83}&\textbf{79.15}&\textbf{82.07}&\textbf{78.03}&\textbf{80.00}\\
    \hline
         {DAPT} \cite{zhang2023dept}, \textit{ICCV23}  &\textbf{81.07}&75.40&78.13&\textbf{78.87}&50.73&61.75&\textbf{83.67}&33.97&48.32&83.80&69.07&75.72\\
    \textbf{+CaPT}(\textit{Ours}) &81.03&\textbf{75.80}&\textbf{78.33}&76.53&\textbf{55.73}&\textbf{64.50}&79.67&\textbf{62.73}&\textbf{70.19}&\textbf{84.97}&\textbf{76.20}&\textbf{80.34}\\
    \hline
         {DePT}, \textit{CVPR24}(\textit{Ours})  &\textbf{82.43}&77.73&80.01&82.30&58.80&68.59&89.13&67.87&77.06&85.77&77.00&81.15\\
     \textbf{+CaPT}(\textit{Ours}) &82.37&\textbf{78.87}&\textbf{80.05}&\textbf{82.70}&\textbf{62.47}&\textbf{71.17}&\textbf{91.37}&\textbf{73.93}&\textbf{81.73}&\textbf{85.83}&\textbf{78.47}&\textbf{81.98}\\
     \hline
    \end{tabular}
  \label{table—B2N}
\end{table*}

\subsection{Evaluation on Flexibility and Effectiveness}
\label{exp.res}
In this section, we adopt the {base-to-new generalization} setting to comprehensively evaluate the flexibility and effectiveness of the proposed Class-adaptive Prompt Tuning (CaPT) approach. Additionally, the performance of CaPT under cross-dataset and cross-domain generalization settings is detailed in Appendix \ref{secA1}.

In line with those baseline approaches, we first split each dataset into two equal sets of classes to create the base and new tasks. 
We then conduct PT on the base task and evaluate the model's performance on testing samples of both the base and new tasks. 
Table \ref{table—B2N} reports the base-to-new generalization performance of the five baseline methods with or without our CaPT framework across 11 datasets.
Analyzing the average results presented in the table, we can notice a tradeoff between accuracies on base tasks and new tasks for most baseline methods. For instance, ProGrad outperforms CoOp on new tasks but lags behind CoOp on base tasks. Remarkably, CaPT enhances the performance of all baselines consistently without compromising accuracy on either base or new tasks. 
Specifically, CaPT boosts each baseline's performance across base-task, new-task, and harmonic-mean ACCs.
However, we also observe some failure instances, e.g., CaPT does not significantly enhance performance on most baseline methods for the OxfordPets dataset. 
Possible explanations are twofold. On the one hand, the optimal hyperparameters for CaPT vary significantly across different datasets and baselines, but we used fixed parameters across all. On the other hand, when there are minimal category shifts between downstream tasks and the upstream data used for model pretraining, the advantages of CaPT as well as existing PT methods for knowledge transferring become less pronounced. Similar conclusions were made in \cite{luo2023closer,zhang2023dept}.

\subsection{Comparison with SOTAs}
\label{exp.sota}
So far, we can see that our proposed CaPT can be flexibly plugged into existing unconditional PT methods to improve their generalization capabilities on both base and new tasks—i.e., CaPT can help tackle the BNT problem for all of them.
It would be interesting to explore whether integrating CaPT with a strong unconditional PT baseline can outperform previous state-of-the-art conditional and unconditional competitors.
To this end, we propose DeCaPT—a new conditional PT method that integrates CaPT with our recently proposed DePT \cite{zhang2023dept} framework, and compare DeCaPT with several state-of-the-art conditional and unconditional PT approaches. 
The obtained base-to-new generalization results of DeCaPT and other state-of-the-art approaches on the 11 datasets are reported in Table \ref{table-sota}, where CoCoOp \cite{zhou2022conditional} and Bayes\cite{derakhshani2023bayesian} are conditional PT approaches and others are unconditional approaches.
For a fair comparison, the results of the competitors are sourced from their respective published papers, except for the results of DAPT \cite{cho2023dapt} and Bayes \cite{derakhshani2023bayesian}, which are reproduced by us.
As observed, DeCaPT establishes new state-of-the-art results by consistently outperforming both conditional and unconditional PT methods in terms of base-task, new-task and harmonic-mean accuracies.
In particular, from the average H ACC over 11 datasets,  DeCaPT outperforms the state-of-the-art conditional approach Bayes \cite{derakhshani2023bayesian} by {{3.49}}\%
Furthermore, we observe that previous conditional PT methods do not demonstrate superiority over unconditional methods. In contrast, DeCaPT exhibits significant advantages by learning class-adaptive prompts using textual class information from intra-task classes.

\begin{table*}[t]
\setlength{\abovecaptionskip}{0.cm}  
\setlength{\belowcaptionskip}{0.1cm}
\tabcolsep 0.042in
    \footnotesize
    \centering
    \caption{Comparison with state-of-the-art unconditional and conditional PT methods on 11 datasets. $\dagger$ indicates conditional PT methods. $*$ reproduced results. The best (second best) average results are \textbf{bold} (\underline{underlined}).}
    \begin{tabular}{l|c|ccccccccc|cc|c}
        \hline
        \multirow{2}{*}{Datasets} & \multirow{2}{*}{ACC} & CoOp & ProDA& ProGrad& KgCoOp &MaPLe  & LFA& DAPT$^{*}$ & RPO& DePT & CoCoOp$^{\dagger}$ & Bayes$^{\dagger*}$  & \textbf{DeCaPT}$^{\dagger}$ \\
        & & \cite{zhou2022coop} & \cite{lu2022prompt}  & \cite{zhu2022prograd}  & \cite{yao2023kgcoop} &\cite{khattak2023maple} & \cite{ouali2023lfa}  & \cite{cho2023dapt}& \cite{lee2023rpo}  & \cite{zhang2023dept} & \cite{zhou2022conditional} & \cite{derakhshani2023bayesian} & {({Ours})} \\
        \hline
        \multirow{3}{*}{ImgNet} & Base & 76.47 & 75.40  &77.02  &75.83 & 76.66 & 76.89  & 76.77 &76.60 &  77.13& 75.98 & 76.00 &  77.00 \\
        & New & 67.88 &70.23  &66.66  &69.96 &70.54  &69.36 & 69.70 & 73.54  &  70.27 &  70.43 & 70.80 &  70.00 \\
        & H & 71.92  & 72.72  &71.46  &72.78 &  73.47& 72.93&73.06 &71.57 & 74.00& 73.10 &73.31 &  73.33\\
        \hline
        \multirow{3}{*}{Caltech} & Base & 98.00  &98.27 &98.02  &97.72 & 97.74 &98.41  & 97.90& 97.97& 98.30 &97.96  &  97.40 & 98.40\\
        & New & 89.81  &93.23 &93.89  &94.39 &94.36  &93.93 &94.37 & 94.37&  94.67 & 93.81  & 94.50 &94.60 \\
        & H & 93.73  &95.68 & 95.91 &96.03 &96.02  &96.13 &96.10 & 96.03 & 96.45&95.84 &   95.93 &96.46 \\
        \hline
        \multirow{3}{*}{OxfPets} & Base & 93.67 &95.43  &95.07  &94.65 & 95.43 & 95.13& 95.57 &94.63 & 94.37 &95.20 & 95.60 &  93.87 \\
        & New & 95.29  &97.83 &97.63  &97.76 & 97.76 &96.23 & 97.47 &97.50  & 97.23 & 97.69  &  97.90 &97.57 \\
        & H & 94.47  & 96.62  &96.33  &96.18 & 96.58 &95.68 & 96.51 &96.05 & 95.78 & 96.43 &96.74 &   95.68 \\
        \hline
        \multirow{3}{*}{StanCars} & Base & 78.12  &74.70  &77.68  &71.76 & 72.94 & 76.32& 72.37 & 73.87 &79.23& 70.49 &  68.80 & 79.87 \\
        & New & 60.40  &71.20 &68.63  &75.04 & 74.00 & 74.88 &74.50 &75.53 &74.57 & 73.59  &  75.10 & 74.77 \\
        & H & 68.13  &72.91 &72.88  &73.36 & 73.57 & 75.59&73.42 & 74.69 &76.83 &72.01  &   71.81 & 77.23 \\
        \hline
        \multirow{3}{*}{Flowers} & Base & 97.60  &97.70 &95.54  &95.00 & 95.92  & 97.34&96.37 &94.13 & 97.97 & 94.87 &  93.10 & 98.53 \\
        & New & 59.67  &68.68 &71.87  &74.73 & 72.46 & 75.44&71.53 & 76.67 &  75.50 &71.75 &   72.60 &75.43 \\
        & H & 74.06  & 80.66 &82.03  &83.65 & 82.56 &85.00  &82.11 & 84.50& 85.28& 81.71  &81.58 & 85.45 \\
        \hline
        \multirow{3}{*}{Food101} & Base & 88.33  & 90.30  &90.37  &90.50 & 90.71 &90.52 &90.23 &90.33 & {90.50} &90.70 &90.80  &  90.47 \\
        & New & 82.26  & 88.57 &89.59  &91.70 &92.05  & 91.48&91.17 & 90.83 &  91.70 &91.29 & 92.10 & {91.63} \\
        & H & 85.19  & 89.43 &98.98   &91.09& 91.38 &91.00 & 90.70& 90.58 &{91.10} & 90.99 & 91.45 &  91.05 \\
        \hline
        \multirow{3}{*}{Aircraft} & Base & 40.44  &36.90  &40.54  &36.21 & 37.44 &41.48 &28.50  &37.33 & 42.90 &33.41 &  34.20  & 43.47 \\
        & New & 22.30  &34.13  &27.57  &33.55 & 35.61 &32.29 &23.13 &34.20  &35.03 & 23.71 & 36.00 &  35.93 \\
        & H & 28.75  & 35.46 &32.82  &34.93 & 36.50 &36.31 &25.65 & 35.70 & 38.59 &27.74 &  35.08 & 39.34 \\
        \hline
        \multirow{3}{*}{SUN397} & Base & 80.60  &78.67 & 81.26 &80.29 & 80.82 &82.13  &81.07 &80.60 & 82.43 &79.74  & 79.00 &  82.37 \\
        & New & 65.89  &  76.93 & 74.17 &76.53 & 78.70 & 77.20 &75.40 & 77.80& 77.73 &76.86 &77.80 &  78.87 \\
        & H & 72.51  & 77.79 &77.55  &78.36 & 79.75 &79.59 &78.13 & 79.18 &80.01 & 78.27  & 78.40 & {80.05} \\
        \hline
        \multirow{3}{*}{DTD} & Base & 79.44  & 80.67 &77.35  &77.55 & 80.36 &81.29 &78.87  &76.70 &   82.30&77.01 & 74.10 &82.70 \\
        & New & 41.18  &56.48 &52.35  &54.99 & 59.18 &60.63 &50.73 & 62.13 & 59.30&56.00 &  58.80 & 62.47 \\
        & H & 54.24  &66.44 &62.45  &64.35 &68.16  & 69.46&61.75 & 68.61&68.59&  64.85  & 65.88 & 71.17 \\
        \hline
        \multirow{3}{*}{EuroSAT} & Base & 92.19 & 83.90 &90.11  &85.64 & 94.07 &93.40 &83.67  &86.63 & 89.13 &87.49  &63.90 &  91.37 \\
        & New & 54.74  & 66.00  & 60.89 & 64.34& 73.23 &71.24  &33.97 &68.97 & 67.87 &60.04 &68.50 &  73.93 \\
        & H & 68.69  &73.88  &72.67  & 73.48& 82.30 &80.83 &48.32  &76.79 &77.06 &71.21 &  66.12 & 81.73 \\
        \hline
        \multirow{3}{*}{UCF101} & Base & 84.69  & 85.23  & 84.33 & 82.89 & 83.00 & 86.97 &83.80 &83.67 & 85.77 & 82.33 &79.70 & 85.83 \\
        & New & 56.05  &  71.97 & 74.94 & 76.67 &78.66  &77.48 & 69.07 &75.43 & 77.00&73.45 & 78.50 &78.47 \\
        & H & 67.46  &  78.04 & 79.35 & 79.65& 80.77 &81.95 & 75.72& 79.34 &81.15& 77.64 & 79.10 & 81.98\\
        \hline
        \rowcolor{green!8} & Base & 82.69  & 81.56 &82.48  &80.73 & 82.28 &83.62 &80.58 & 81.13 & \underline{83.64} & 80.47 & 77.51 & \textbf{83.99} \\
        \rowcolor{green!8}\textbf{Average}& New & 63.22  & 72.30 &70.75  &73.60 & \underline{75.14} &74.56 & 66.23 & 75.00 & 74.58 &71.69 &74.83 &  \textbf{75.70} \\
        \rowcolor{green!8}& H & 71.66  & 76.65  & 76.16 & 77.00& 78.55 &78.83 &72.71 &77.78  & \underline{78.85} & 75.83 &76.14 & \textbf{79.63} \\
        \hline
    \end{tabular}
    \label{table-sota}
\end{table*}

\begin{table}[tp]
\setlength{\abovecaptionskip}{-0.4cm}  
\setlength{\belowcaptionskip}{-0.cm} 
		\centering
		\tabcolsep 0.0018in
		\footnotesize
        \caption{Ablation study on the devised components.  {\textcircled{1}}: unconditional PT,  {\textcircled{2}}: conditional PT with Textual Class Information (TCI), {\textcircled{3}} (\textbf{CaPT}): conditional PT with TCI and the margin-ITM loss.
        }
        \begin{tabular}{c|cc|ccc}
		      \hline
			 \multirow{2}{*}{ID} &  \multirow{2}{*}{TCI} &  \multirow{2}{*}{margin-ITM}  & \multicolumn{3}{c}{Avg ACC (\%)} \\
			\cline{4-6}
			  &  &  & Base & New & H \\
			\hline
			 {\textcircled{1}} & \textcolor{gray!50}{\XSolidBrush} &  \textcolor{gray!50}{\XSolidBrush} & 80.82   & 69.68 & 74.84 \\
			 {\textcircled{2}} & \Checkmark & \textcolor{gray!50}{\XSolidBrush}  & \textbf{81.35}(\textcolor{red}{+0.53}) & 72.94(\textcolor{red}{+3.26}) &76.91(\textcolor{red}{+2.07})  \\
   		\rowcolor{green!8}{\textcircled{3}}&\Checkmark  & \Checkmark & 81.32(\textcolor{red}{+0.50})  & \textbf{73.92}(\textcolor{red}{+4.24}) & \textbf{77.44}(\textcolor{red}{+2.60})  \\
            \hline
		\end{tabular}
		\label{table.ablation}
\end{table}

\begin{table}[tp]
\setlength{\abovecaptionskip}{-0.4cm}  
\setlength{\belowcaptionskip}{-0.cm} 
		\centering
		 \tabcolsep 0.09in
		\footnotesize
        \caption{Ablation study on the Word2Vec model $\varphi(\cdot)$.}
        \begin{tabular}{c|c|ccc}
			\hline
			 \multirow{2}{*}{ID} &  \multirow{2}{*}{$\varphi(\cdot)$}  & \multicolumn{3}{c}{Avg ACC (\%)} \\
			\cline{3-5}
			  &    & Base & New & H \\
			\hline
			 {\textcircled{1}} & Fasttext \cite{joulin2016fasttext}  & 80.20   & 71.25 & 75.46 \\
			 {\textcircled{2}} & Glove \cite{pennington2014glove}  & \textbf{81.75}   & 71.46  &76.26  \\
    	   {\textcircled{3}} & Wikipedia2vec \cite{yamada2020wikipedia2vec}   & 81.23 & 71.89  &76.27   \\
   		\rowcolor{green!8}{\textcircled{4}}& CLIP text encoder \cite{gao2024clip} & 81.32   & \textbf{73.92}   & \textbf{77.44}   \\
            \hline
		\end{tabular}
		\label{table.token}
\end{table}

\begin{table}[tp]
\setlength{\abovecaptionskip}{-0.4cm}  
\setlength{\belowcaptionskip}{0.cm} 
		\centering
		 \tabcolsep 0.15in
		\footnotesize
        \caption{Ablation study on the Meta-Net architecture.}
        \begin{tabular}{c|c|ccc}
			\hline
			 \multirow{2}{*}{ID} &  \multirow{2}{*}{Meta-Net}  & \multicolumn{3}{c}{Avg ACC (\%)} \\
			\cline{3-5}
			  &    & Base & New & H \\
			\hline
			 {\textcircled{1}} & Attention  & 78.97   & 72.19 & 75.43 \\
			 {\textcircled{2}} &  Linear   & 81.00  & 72.14  &76.31   \\
   		\rowcolor{green!8}{\textcircled{3}}& MLP & \textbf{81.32}   & \textbf{73.92}   & \textbf{77.44}   \\
            \hline
		\end{tabular}
		\label{table.metanet}
\end{table}

\subsubsection{Ablation Studies}
\label{ablation}
In this section, we conduct ablative studies to further scrutinize CaPT. 
The baseline is CoOp \cite{zhou2022coop}, the average results on 11 datasets are reported.

\keypoint{Effectiveness of the Designed Components.}
The proposed CaPT framework consists of the following components:
1) The textual class information (TCI) conditioned PT mechanism, denoted as TCI. 
2) An margin-adjusted Image-Text Matching loss, denoted as margin-ITM.
Here, we conduct component-wise analysis to scrutinize the impact of each component by alternatively adding one of them to the baseline method. From the obtained results in Table \ref{table.ablation}, we have the following observations.
{{\textit{Firstly}}}, each component in CaPT contributes to the performance improvement.
{{\textit{Secondly}}}, by combining all those components, CaPT improves the baseline method by a large margin, further proving the effectiveness of our CaPT on overcoming the BNT problem for existing PT methods.
{{\textit{Thirdly}}}, the introduction of margin-ITM does not significantly degrade the model’s generalization ability on the base task, while it considerably boosts performance on new tasks.

\keypoint{Impact of the Word2Vec Model $\varphi(\cdot)$.}
In our CaPT, we use a pretrained Word2Vec model $\varphi(\cdot)$ to extract the textual class information (TCI) for both base and new tasks. Here, we conduct experiments to study the impacts of four Word2Vec models on our CaPT, specifically including Fasttext \cite{joulin2016fasttext}, Glove \cite{pennington2014glove}, Wikepedia2vec \cite{yamada2020wikipedia2vec} and the text encoder of the pretrained CLIP model \cite{gao2024clip}. 
The obtained results are shown in Table \ref{table.token}.
\textit{Firstly}, all those results of CaPT in the table outperforms that of the baseline in Table \ref{table.ablation}, demonstrating the effectiveness of the Word2Vec model $\varphi(\cdot)$ for CaPT.
\textit{Secondly}, CaPT achieves comparable results when $\zeta(\cdot)$=Glove and $\zeta(\cdot)$=Wikepedia2vec.
\textit{Thirdly}, CaPT obtains the best Base ACC when $\zeta(\cdot)$=Glove, and establishes the best H ACC when $\zeta(\cdot)$= Text encoder of CLIP.
Thus, we adopt the pretrained text encoder of CLIP to produce the TCI for the base task and new tasks.

\keypoint{Impact of the Meta-Net Architecture.}
The devised CaPT adopts a learnable lightweight network, coined Meta-Net, to obtain class-adaptive features of the extracted TCI of classes. 
Here, we conduct experiments to compare the performance of three different structures of Meta-Net, including a Attention structure (i.e. one self-attention layer), a Linear transformation layer, and a two-layer MLP (the bottleneck dimension is set to 32).
As can be observed in Table \ref{table.metanet}, all those results of CaPT in the table outperforms that of the baseline in Table \ref{table.ablation}, demonstrating the effectiveness of the Meta-Net for learning class-adaptive features for base and new classes.
Besides, MLP consistently outperforms the two other competitors (i.e., Attention, Linear) in terms of base-task, new-task and harmonic-mean accuracies.
Hence, we designs the Meta-Net of CaPT as an MLP to extract class-adaptive features for both the base task and new classes in our experiments. 

\begin{figure}[t]
\setlength{\abovecaptionskip}{-0.4cm}  
\setlength{\belowcaptionskip}{-0.2cm} 
\centering
\includegraphics[width=1.0\linewidth]{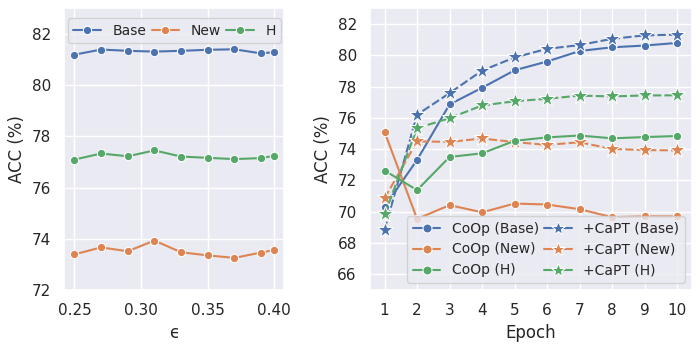} 
\caption{\textbf{Left}: Impact of the scaling weight $\epsilon$ on CaPT. 
\textbf{Right}: Performance of CaPT at different training epochs. The results are the average over 11 datasets.} 
\label{hyper}
\end{figure}

\keypoint{Impact of the Scaling Weight.}
In the proposed margin-ITM loss of CaPT, a scaling weight $\epsilon$ is introduced to scaling the similarity of each pair of base classes.
It is necessary to scrutinize the impact of $\epsilon$ on the performance of CaPT.
For achieving this goal, we respectively set $\epsilon$ to $\{0.25, 0.27, 0.29, 0.31, 0.33, 0.35, 0.37, 0.39\}$, and report the average testing results on the 11 datasets in Fig. \ref{hyper} (\textbf{Left}).
In general, our CaPT is not sensitive to the change of  $\epsilon$ within a certain range (from 0.27 to 0.39).
Particularly, the achieved base-task, new-task and harmonic-mean
accuracies of our CaPT gradually increase as the $\epsilon$ value grows from 0.25 to 0.31, after which the performance gradually decreases. DePT establishes the best results on both base and new tasks when $\epsilon$=0.31, we thus set $\epsilon$=0.31 in our experiments.

\begin{table}[t]
\setlength{\abovecaptionskip}{0.cm}  
\setlength{\belowcaptionskip}{0.1cm} 
		\centering
		 \tabcolsep 0.015in
		\footnotesize
  		\caption{{Computational cost of our CaPT}. “ms”: millisecond per image.}
		\begin{tabular}{l|ccc|c}
		\hline
		  Method  & Train.time & Infer.time & Mem.cost & Avg H (\%) \\
            \hline
            CoOp \cite{zhou2022coop} & 6161s &2.78ms &11267MB  &74.84\\
               +\textbf{DePT}& \color{red}{+4s}\color{black} &\color{red}{+0.05ms}\color{black} &\color{red}+512k\color{black} &\textbf{77.44} (\color{red}{+2.60}\color{black})\\
            \hline            
            KgCoOp \cite{yao2023kgcoop} &5986s &2.73ms &11257MB & 76.98\\
               +\textbf{DePT}  &\color{red}+3s\color{black} &\color{red}+0.01ms\color{black} &\color{red}+512k\color{black} &\textbf{77.83} (\color{red}{+0.85}\color{black})\\
            \hline            
            ProGrad \cite{zhu2022prograd}&9781s &2.82ms &11735MB &76.18\\
             +\textbf{DePT} &\color{red}+4s\color{black} &\color{red}+0.01ms\color{black} &\color{red}+512k\color{black} &\textbf{77.75} (\color{red}{+1.57}\color{black})\\
            \hline            
            DAPT \cite{cho2023dapt} &6313s &2.86ms &11799MB &72.71\\
             +\textbf{DePT} &\color{red}+15s\color{black} &\color{red}+0.02ms\color{black} &\color{red}+512k\color{black} &\textbf{76.60} (\color{red}{+3.89}\color{black})\\
            \hline            
            DePT \cite{zhang2023dept} &6003s &2.74ms &11045MB &78.85\\
             +\textbf{DePT} &\color{red}+8s\color{black} &\color{red}+0.03ms\color{black} &\color{red}+512k\color{black} &\textbf{79.63} (\color{red}{+0.78}\color{black})\\
            \hline
		\end{tabular}
		\label{cost}
\end{table}

\begin{figure*}
\setlength{\abovecaptionskip}{-0.2cm}  
\setlength{\belowcaptionskip}{-0.2cm} 
	\centering	
	\includegraphics[width=0.93\linewidth]{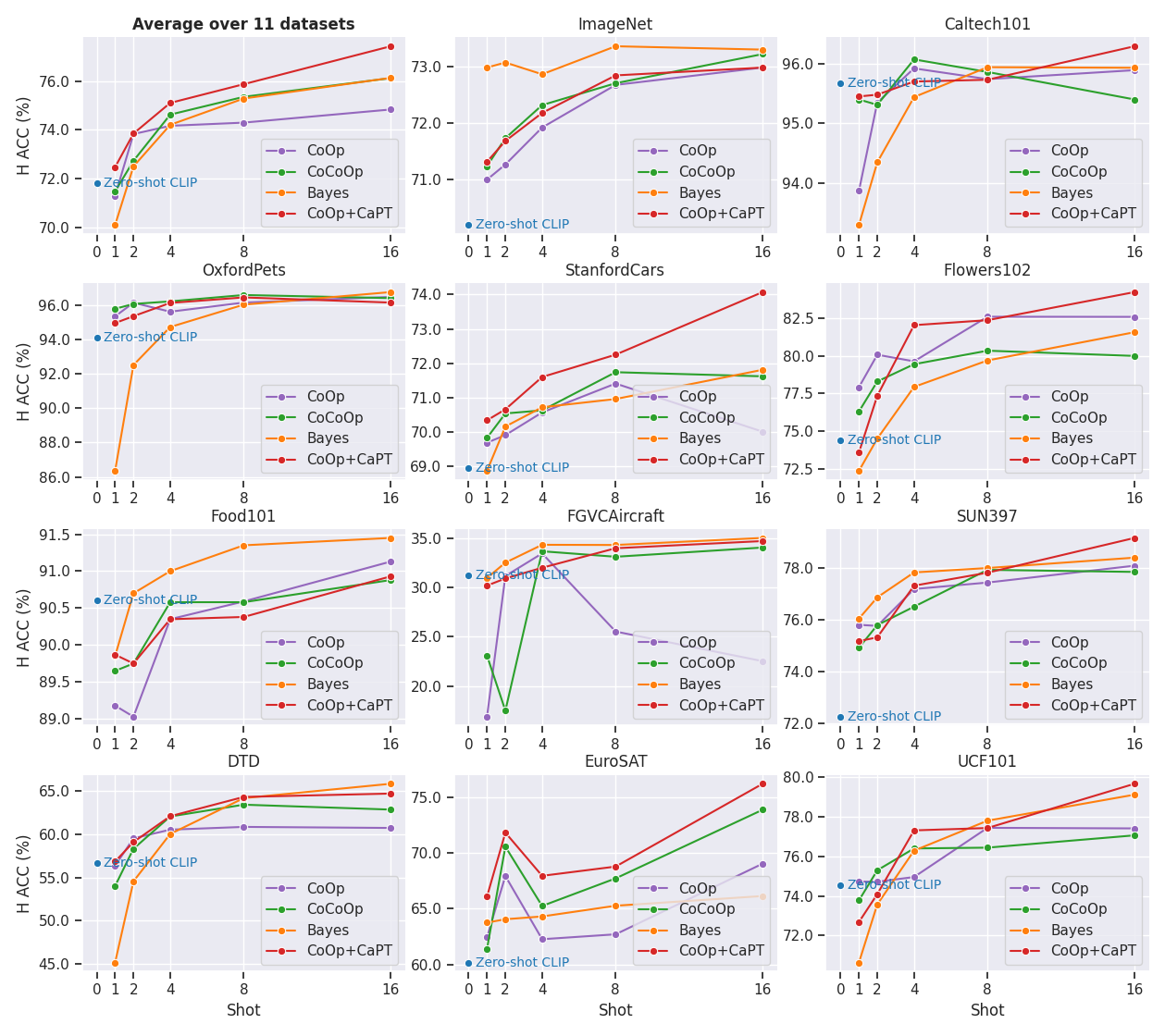} 
	\caption{Effectiveness of our proposed CaPT under different shots. 
        CoCoOp\cite{zhou2022conditional}, Bayes\cite{derakhshani2023bayesian} as well as our CoOp+CaPT are conditional PT paradigms that leverage the same unconditional PT scheme CoOp \cite{zhou2022coop} as baseline.}
	\label{fig.shot}
\end{figure*}

\keypoint{Performance of CaPT at Different Epochs.}
In Fig. \ref{hyper} (\textbf{Right}), we report the obtained results of the baseline method with or without our CaPT at different training epochs.
As can be observed, our CaPT consistently improves the baseline method after epoch 2, in terms of base-task, new-task, and harmonic mean (H) accuracies. 
One possible reason for the failure case at epoch 1 is that the parameters of the MLP network are initialized randomly, thus it is difficult to fully capture class-adaptive features of classes with only one training epoch.
We also see the baseline fails to address the BNT problem during PT—overall, the H ACC of the baseline decreases as the Base ACC increases from epoch 1 to epoch 10. 
It is evident that CaPT effectively mitigates the BNT problem. 
The H ACC of both the baseline method and our CaPT saturates at epoch 10.

\keypoint{Performance of CaPT with Different Shots.} 
In our experiments, we follow the five baseline methods to evaluate the generalization performance of the tuned models on many-way 16-shot base tasks, i.e, only 16 training examples of each base class are used to learn task-specific prompts.
It is interesting to investigate the effectiveness of CaPT with different shots.
Fig. \ref{fig.shot} reports the base-to-new generalization performance of CoOp\cite{zhou2022coop}, CoCoOp\cite{zhou2022conditional}, Bayes\cite{derakhshani2023bayesian} and CoOp+CaPT with different shots.
From the average results on 11 datasets, we see that CaPT consistently improves the baseline method (CoOp) and outperforms the two conditional PT approaches CoCoOp and Bayes across 1-shot, 2-shot 4-shot, 8-shot and 16-shot settings, further validating the flexibility and effectiveness of our CaPT.
It is also worth noting that except for our CaPT, the average 1-shot classification ACCs of CoOp, CoCoOp and Bayes are inferior to the obtained zero-shot generalization results of CLIP. 
This suggests that our proposed CaPT is more robust to data-scarce scenarios compared to those methods.

\subsection{Computational Cost}
So far, by performing extensive experiments using five baseline methods and 11 datasets, we have already demonstrated the strong flexibility and effectiveness of our proposed CaPT (as well as the devised components) in tackling the BNT problem for existing unconditional PT methods.
This raises a key question: how much additional computational cost does CaPT incur while achieving such performance improvements on these methods?
To answer the question, we report the training time (seconds), inference time (millisecond per image, ms), memory cost (MB) and the obtained H ACC of the five baseline methods with or without our CaPT in Table \ref{cost}, where the H ACCs are the average over the 11 datasets. All experiments are performed on a single NVIDIA V100 GPU. 
As can be seen, the additional computational cost of CaPT is low (even {negligible}) compared to the performance improvement over those baseline methods, showing the high efficiency of CaPT. 
Combined with the previous experimental results, we can see that our proposed CaPT is simple, flexible, effective and efficient.

\section{Conclusions}
In this work, we first identify an overlooked problem of existing conditional PT approaches: leveraging VII as the “condition” of prompts yields suboptimal conditional PT performance, and even random noise-conditioned prompts can outperform the VII-conditioned counterparts.
Through further analysis, we reveal that the key to addressing the Base-New Tradeoff (BNT) problem in PT is to learn dynamic prompts conditioned on Textual Class Information (TCI). 
Motivated by this, we propose Class-adaptive Prompt Tuning (CaPT), which facilitates the rapid adaptation of tuned models to new classes by leveraging TCI-conditioned prompts derived from base classes. 
Notably, CaPT can be integrated as a plugin to overcome the BNT issue for existing unconditional PT approaches with minimal additional computational overhead. 
Extensive experiments across 11 datasets validate the flexibility and effectiveness of CaPT.
Our code is publicly available at \url{https://github.com/Koorye/CaPT}.

\bibliography{Rethinking}

\begin{thebibliography}{10}
\providecommand{\doi}[1]{\url{https://doi.org/#1}}
\bibcommenthead

\bibitem[\protect\citeauthoryear{Zhang et~al.}{2024}]{zhang2023dept}
Zhang J, Wu S, Gao L, Shen H, Song J.
\newblock Dept: Decoupled prompt tuning.
\newblock CVPR. 2024;.

\bibitem[\protect\citeauthoryear{Radford et~al.}{2021}]{radford2021clip}
Radford A, Kim JW, Hallacy C, Ramesh A, Goh G, Agarwal S, et~al.
\newblock Learning transferable visual models from natural language supervision.
\newblock In: ICML; 2021. p. 8748--8763.

\bibitem[\protect\citeauthoryear{Luo et~al.}{2023}]{luo2023closer}
Luo X, Wu H, Zhang J, Gao L, Xu J, Song J.
\newblock A closer look at few-shot classification again.
\newblock In: ICML. PMLR; 2023. p. 23103--23123.

\bibitem[\protect\citeauthoryear{Derakhshani et~al.}{2023}]{derakhshani2023bayesian}
Derakhshani MM, Sanchez E, Bulat A, da~Costa VGT, Snoek CG, Tzimiropoulos G, et~al.
\newblock Bayesian prompt learning for image-language model generalization.
\newblock In: CVPR; 2023. p. 15237--15246.

\bibitem[\protect\citeauthoryear{Khattak et~al.}{2023}]{khattak2023maple}
Khattak MU, Rasheed H, Maaz M, Khan S, Khan FS.
\newblock Maple: Multi-modal prompt learning.
\newblock In: CVPR; 2023. p. 19113--19122.

\bibitem[\protect\citeauthoryear{Zhou et~al.}{2022}]{zhou2022conditional}
Zhou K, Yang J, Loy CC, et al.
\newblock Conditional prompt learning for vision-language models.
\newblock In: CVPR; 2022. p. 16816--16825.

\bibitem[\protect\citeauthoryear{Zhou et~al.}{2022}]{zhou2022learning}
Zhou K, Yang J, Loy CC, Liu Z.
\newblock Learning to prompt for vision-language models.
\newblock International Journal of Computer Vision. 2022;130(9):2337--2348.

\bibitem[\protect\citeauthoryear{Yao et~al.}{2023}]{yao2023kgcoop}
Yao H, Zhang R, Xu C.
\newblock Visual-language prompt tuning with knowledge-guided context optimization.
\newblock In: CVPR; 2023. p. 6757--6767.

\bibitem[\protect\citeauthoryear{Zhu et~al.}{2023}]{zhu2022prograd}
Zhu B, Niu Y, Han Y, Wu Y, Zhang H.
\newblock Prompt-aligned gradient for prompt tuning.
\newblock ICCV. 2023;.

\bibitem[\protect\citeauthoryear{Cho et~al.}{2023}]{cho2023dapt}
Cho E, Kim J, Kim HJ.
\newblock Distribution-aware prompt tuning for vision-language models.
\newblock In: ICCV; 2023. p. 22004--22013.

\bibitem[\protect\citeauthoryear{Jin et~al.}{2023}]{jin2023context}
Jin Z, Hayat M, Yang Y, Guo Y, Lei Y.
\newblock Context-aware Alignment and Mutual Masking for {3D}-Language Pre-training.
\newblock In: CVPR; 2023. p. 10984--10994.

\bibitem[\protect\citeauthoryear{Min et~al.}{2023}]{min2023recent}
Min B, Ross H, Sulem E, Veyseh APB, Nguyen TH, Sainz O, et~al.
\newblock Recent advances in natural language processing via large pre-trained language models: A survey.
\newblock ACM Computing Surveys. 2023;56(2):1--40.

\bibitem[\protect\citeauthoryear{Khurana et~al.}{2023}]{khurana2023natural}
Khurana D, Koli A, Khatter K, Singh S.
\newblock Natural language processing: State of the art, current trends and challenges.
\newblock Multimedia tools and applications. 2023;82(3):3713--3744.

\bibitem[\protect\citeauthoryear{Kim et~al.}{2021}]{kim2021vilt}
Kim W, Son B, Kim I.
\newblock Vilt: Vision-and-language transformer without convolution or region supervision.
\newblock In: ICML. PMLR; 2021. p. 5583--5594.

\bibitem[\protect\citeauthoryear{Jia et~al.}{2021}]{jia2021scaling}
Jia C, Yang Y, Xia Y, Chen YT, Parekh Z, Pham H, et~al.
\newblock Scaling up visual and vision-language representation learning with noisy text supervision.
\newblock In: ICML; 2021. p. 4904--4916.

\bibitem[\protect\citeauthoryear{Huo et~al.}{2021}]{huo2021wenlan}
Huo Y, Zhang M, Liu G, Lu H, Gao Y, Yang G, et~al.
\newblock WenLan: Bridging vision and language by large-scale multi-modal pre-training.
\newblock arXiv preprint arXiv:210306561. 2021;.

\bibitem[\protect\citeauthoryear{Li et~al.}{2021}]{li2021align}
Li J, Selvaraju R, Gotmare A, Joty S, Xiong C, Hoi SCH.
\newblock Align before fuse: Vision and language representation learning with momentum distillation.
\newblock NeurIPS. 2021;34:9694--9705.

\bibitem[\protect\citeauthoryear{Mo et~al.}{2023}]{DMG_ICML}
Mo Y, Lei Y, Shen J, Shi X, Shen HT, Zhu X.
\newblock Disentangled Multiplex Graph Representation Learning.
\newblock In: ICML. vol. 202; 2023. p. 24983--25005.

\bibitem[\protect\citeauthoryear{Lu et~al.}{2019}]{lu2019vilbert}
Lu J, Batra D, Parikh D, Lee S.
\newblock Vilbert: Pretraining task-agnostic visiolinguistic representations for vision-and-language tasks.
\newblock NeurIPS. 2019;32.

\bibitem[\protect\citeauthoryear{Ma et~al.}{2022}]{ma2022x}
Ma Y, Xu G, Sun X, Yan M, Zhang J, Ji R.
\newblock X-clip: End-to-end multi-grained contrastive learning for video-text retrieval.
\newblock In: ACM MM; 2022. p. 638--647.

\bibitem[\protect\citeauthoryear{Zhu et~al.}{2023}]{zhu2023complementarity}
Zhu J, Zeng P, Gao L, Li G, Liao D, Song J.
\newblock Complementarity-aware space learning for video-text retrieval.
\newblock IEEE Transactions on Circuits and Systems for Video Technology. 2023;.

\bibitem[\protect\citeauthoryear{Rao et~al.}{2022}]{rao2022denseclip}
Rao Y, Zhao W, Chen G, Tang Y, Zhu Z, Huang G, et~al.
\newblock Denseclip: Language-guided dense prediction with context-aware prompting.
\newblock In: CVPR; 2022. p. 18082--18091.

\bibitem[\protect\citeauthoryear{Zhou et~al.}{2022}]{zhou2022extract}
Zhou C, Loy CC, Dai B.
\newblock Extract free dense labels from clip.
\newblock In: ECCV. Springer; 2022. p. 696--712.

\bibitem[\protect\citeauthoryear{Wang et~al.}{2022}]{wang2022clip}
Wang C, Chai M, He M, Chen D, Liao J.
\newblock Clip-nerf: Text-and-image driven manipulation of neural radiance fields.
\newblock In: CVPR; 2022. p. 3835--3844.

\bibitem[\protect\citeauthoryear{Patashnik et~al.}{2021}]{patashnik2021styleclip}
Patashnik O, Wu Z, Shechtman E, Cohen-Or D, Lischinski D.
\newblock Styleclip: Text-driven manipulation of stylegan imagery.
\newblock In: CVPR; 2021. p. 2085--2094.

\bibitem[\protect\citeauthoryear{Houlsby et~al.}{2019}]{houlsby2019adapters}
Houlsby N, Giurgiu A, Jastrzebski S, Morrone B, De~Laroussilhe Q, Gesmundo A, et~al.
\newblock Parameter-efficient transfer learning for NLP.
\newblock In: ICML; 2019. p. 2790--2799.

\bibitem[\protect\citeauthoryear{Triantafillou et~al.}{2021}]{triantafillou2021learning}
Triantafillou E, Larochelle H, Zemel R, Dumoulin V.
\newblock Learning a universal template for few-shot dataset generalization.
\newblock In: ICML; 2021. p. 10424--10433.

\bibitem[\protect\citeauthoryear{Zhang et~al.}{2025}]{zhang2025reliable}
Zhang J, Song J, Gao L, Sebe N, Shen HT.
\newblock Reliable Few-shot Learning under Dual Noises.
\newblock arXiv preprint arXiv:250616330. 2025;.

\bibitem[\protect\citeauthoryear{Li and Liang}{2021}]{li2021prefix}
Li XL, Liang P.
\newblock Prefix-Tuning: Optimizing Continuous Prompts for Generation.
\newblock In: ACL; 2021. p. 4582--4597.

\bibitem[\protect\citeauthoryear{Wu et~al.}{2025}]{wu2025skip}
Wu S, Zhang J, Zeng P, Gao L, Song J, Shen HT.
\newblock Skip Tuning: Pre-trained Vision-Language Models are Effective and Efficient Adapters Themselves.
\newblock In: CVPR; 2025. p. 14723--14732.

\bibitem[\protect\citeauthoryear{Hu et~al.}{2021}]{hu2021lora}
Hu EJ, Shen Y, Wallis P, Allen-Zhu Z, Li Y, Wang S, et~al.
\newblock Lora: Low-rank adaptation of large language models.
\newblock arXiv preprint arXiv:210609685. 2021;.

\bibitem[\protect\citeauthoryear{Wang et~al.}{2022}]{wang2022defo}
Wang F, Li M, Lin X, Lv H, Schwing AG, Ji H.
\newblock Learning to decompose visual features with latent textual prompts.
\newblock ICLR. 2022;.

\bibitem[\protect\citeauthoryear{Chen et~al.}{2022}]{chen2022prompt}
Chen G, Yao W, Song X, Li X, Rao Y, Zhang K.
\newblock Prompt learning with optimal transport for vision-language models.
\newblock ICLR. 2022;.

\bibitem[\protect\citeauthoryear{Yang et~al.}{2025}]{yang2025consistent}
Yang M, Yin J, Gu Y, Deng C, Zhang H, Zhu H.
\newblock Consistent Prompt Tuning for Generalized Category Discovery.
\newblock International Journal of Computer Vision. 2025;p. 1--28.

\bibitem[\protect\citeauthoryear{Xu et~al.}{2025}]{xu2025progressive}
Xu C, Zhu Y, Shen H, Chen B, Liao Y, Chen X, et~al.
\newblock Progressive visual prompt learning with contrastive feature re-formation.
\newblock International Journal of Computer Vision. 2025;133(2):511--526.

\bibitem[\protect\citeauthoryear{Wu et~al.}{2024}]{wu2024hybridprompt}
Wu J, Zhang T, Zhang Y.
\newblock HybridPrompt: Domain-Aware Prompting for Cross-Domain Few-Shot Learning.
\newblock International Journal of Computer Vision. 2024;132(12):5681--5697.

\bibitem[\protect\citeauthoryear{Zhang et~al.}{2024}]{zhang2023deta}
Zhang J, Gao L, Luo X, Shen H, Song J.
\newblock DETA: Denoised Task Adaptation for Few-Shot Learning.
\newblock CVPR. 2024;.

\bibitem[\protect\citeauthoryear{Fu et~al.}{2021}]{fu2021meta}
Fu Y, Fu Y, Jiang YG.
\newblock Meta-fdmixup: Cross-domain few-shot learning guided by labeled target data.
\newblock In: ACM MM; 2021. p. 5326--5334.

\bibitem[\protect\citeauthoryear{Fu et~al.}{2023}]{fu2023styleadv}
Fu Y, Xie Y, Fu Y, Jiang YG.
\newblock StyleAdv: Meta Style Adversarial Training for Cross-Domain Few-Shot Learning.
\newblock In: CVPR; 2023. p. 24575--24584.

\bibitem[\protect\citeauthoryear{Zhang et~al.}{2022}]{zhang2022tip}
Zhang R, Zhang W, Fang R, Gao P, Li K, Dai J, et~al.
\newblock Tip-adapter: Training-free adaption of clip for few-shot classification.
\newblock In: ECCV. Springer; 2022. p. 493--510.

\bibitem[\protect\citeauthoryear{Zhang et~al.}{2023}]{zhang2023prompt}
Zhang R, Hu X, Li B, Huang S, Deng H, Qiao Y, et~al.
\newblock Prompt, generate, then cache: Cascade of foundation models makes strong few-shot learners.
\newblock In: CVPR; 2023. p. 15211--15222.

\bibitem[\protect\citeauthoryear{Jia et~al.}{2022}]{jia2022vpt}
Jia M, Tang L, Chen BC, Cardie C, Belongie S, Hariharan B, et~al.
\newblock Visual prompt tuning.
\newblock In: ECCV; 2022. p. 709--727.

\bibitem[\protect\citeauthoryear{Huang et~al.}{2023}]{huang2023diversity}
Huang Q, Dong X, Chen D, Zhang W, Wang F, Hua G, et~al.
\newblock Diversity-Aware Meta Visual Prompting.
\newblock In: CVPR; 2023. p. 10878--10887.

\bibitem[\protect\citeauthoryear{Khattak et~al.}{2023}]{Khattak_2023_ICCV}
Khattak MU, Wasim ST, Naseer M, Khan S, Yang MH, Khan FS.
\newblock Self-regulating Prompts: Foundational Model Adaptation without Forgetting.
\newblock In: ICCV; 2023. p. 15190--15200.

\bibitem[\protect\citeauthoryear{Zhou et~al.}{2022}]{zhou2022coop}
Zhou K, Yang J, Loy CC, Liu Z.
\newblock Learning to prompt for vision-language models.
\newblock International Journal of Computer Vision. 2022;130(9):2337--2348.

\bibitem[\protect\citeauthoryear{Joulin et~al.}{2016}]{joulin2016fasttext}
Joulin A, Grave E, Bojanowski P, Douze M, J{\'e}gou H, Mikolov T.
\newblock FastText.zip: Compressing text classification models.
\newblock arXiv preprint arXiv:161203651. 2016;.

\bibitem[\protect\citeauthoryear{Yamada et~al.}{2020}]{yamada2020wikipedia2vec}
Yamada I, Asai A, Sakuma J, Shindo H, Takeda H, Takefuji Y, et~al.
\newblock {W}ikipedia2{V}ec: An Efficient Toolkit for Learning and Visualizing the Embeddings of Words and Entities from {W}ikipedia.
\newblock In: EMNLP. ACL; 2020. p. 23--30.

\bibitem[\protect\citeauthoryear{Pennington et~al.}{2014}]{pennington2014glove}
Pennington J, Socher R, Manning CD.
\newblock Glove: Global vectors for word representation.
\newblock In: EMNLP; 2014. p. 1532--1543.

\bibitem[\protect\citeauthoryear{Gao et~al.}{2024}]{gao2024clip}
Gao P, Geng S, Zhang R, Ma T, Fang R, Zhang Y, et~al.
\newblock Clip-adapter: Better vision-language models with feature adapters.
\newblock International Journal of Computer Vision. 2024;132(2):581--595.

\bibitem[\protect\citeauthoryear{Deng et~al.}{2009}]{deng2009imagenet}
Deng J, Dong W, Socher R, Li LJ, Li K, Fei-Fei L.
\newblock Imagenet: A large-scale hierarchical image database.
\newblock In: CVPR. Ieee; 2009. p. 248--255.

\bibitem[\protect\citeauthoryear{Fei-Fei et~al.}{2004}]{fei2004learning}
Fei-Fei L, Fergus R, Perona P.
\newblock Learning generative visual models from few training examples: An incremental bayesian approach tested on 101 object categories.
\newblock In: CVPRW. IEEE; 2004. p. 178--178.

\bibitem[\protect\citeauthoryear{Parkhi et~al.}{2012}]{parkhi2012cats}
Parkhi OM, Vedaldi A, Zisserman A, Jawahar C.
\newblock Cats and dogs.
\newblock In: CVPR. IEEE; 2012. p. 3498--3505.

\bibitem[\protect\citeauthoryear{Krause et~al.}{2013}]{krause20133d}
Krause J, Stark M, Deng J, Fei-Fei L.
\newblock 3d object representations for fine-grained categorization.
\newblock In: ICCVW; 2013. p. 554--561.

\bibitem[\protect\citeauthoryear{Nilsback and Zisserman}{2008}]{nilsback2008automated}
Nilsback ME, Zisserman A.
\newblock Automated flower classification over a large number of classes.
\newblock In: ICVGIP. IEEE; 2008. p. 722--729.

\bibitem[\protect\citeauthoryear{Bossard et~al.}{2014}]{bossard2014food}
Bossard L, Guillaumin M, Van~Gool L.
\newblock Food-101--mining discriminative components with random forests.
\newblock In: ECCV. Springer; 2014. p. 446--461.

\bibitem[\protect\citeauthoryear{Maji et~al.}{2013}]{maji2013fine}
Maji S, Rahtu E, Kannala J, Blaschko M, Vedaldi A.
\newblock Fine-grained visual classification of aircraft.
\newblock arXiv preprint arXiv:13065151. 2013;.

\bibitem[\protect\citeauthoryear{Helber et~al.}{2019}]{helber2019eurosat}
Helber P, Bischke B, Dengel A, Borth D.
\newblock Eurosat: A novel dataset and deep learning benchmark for land use and land cover classification.
\newblock IEEE Journal of Selected Topics in Applied Earth Observations and Remote Sensing. 2019;12(7):2217--2226.

\bibitem[\protect\citeauthoryear{Soomro et~al.}{2012}]{soomro2012ucf101}
Soomro K, Zamir AR, Shah M.
\newblock UCF101: A dataset of 101 human actions classes from videos in the wild.
\newblock arXiv preprint arXiv:12120402. 2012;.

\bibitem[\protect\citeauthoryear{Cimpoi et~al.}{2014}]{cimpoi2014describing}
Cimpoi M, Maji S, Kokkinos I, Mohamed S, Vedaldi A.
\newblock Describing textures in the wild.
\newblock In: CVPR; 2014. p. 3606--3613.

\bibitem[\protect\citeauthoryear{Xiao et~al.}{2010}]{xiao2010sun}
Xiao J, Hays J, Ehinger KA, Oliva A, Torralba A.
\newblock Sun database: Large-scale scene recognition from abbey to zoo.
\newblock In: CVPR. IEEE; 2010. p. 3485--3492.

\bibitem[\protect\citeauthoryear{Recht et~al.}{2019}]{recht2019imagenet}
Recht B, Roelofs R, Schmidt L, Shankar V.
\newblock Do imagenet classifiers generalize to imagenet?
\newblock In: ICML. PMLR; 2019. p. 5389--5400.

\bibitem[\protect\citeauthoryear{Wang et~al.}{2019}]{wang2019learning}
Wang H, Ge S, Lipton Z, Xing EP.
\newblock Learning robust global representations by penalizing local predictive power.
\newblock NeurIPS. 2019;32.

\bibitem[\protect\citeauthoryear{Gao et~al.}{2022}]{gao2022generating}
Gao H, Zhang H, Yang X, Li W, Gao F, Wen Q.
\newblock Generating natural adversarial examples with universal perturbations for text classification.
\newblock Neurocomputing. 2022;471:175--182.

\bibitem[\protect\citeauthoryear{Hendrycks et~al.}{2021}]{hendrycks2021many}
Hendrycks D, Basart S, Mu N, Kadavath S, Wang F, Dorundo E, et~al.
\newblock The many faces of robustness: A critical analysis of out-of-distribution generalization.
\newblock In: ICCV; 2021. p. 8340--8349.

\bibitem[\protect\citeauthoryear{Khattak et~al.}{2023}]{khattak2023promptsrc}
Khattak MU, Wasim ST, Naseer M, Khan S, Yang MH, Khan FS.
\newblock Self-regulating prompts: Foundational model adaptation without forgetting.
\newblock In: ICCV; 2023. p. 15190--15200.

\bibitem[\protect\citeauthoryear{Lu et~al.}{2022}]{lu2022prompt}
Lu Y, Liu J, Zhang Y, Liu Y, Tian X.
\newblock Prompt distribution learning.
\newblock In: CVPR; 2022. p. 5206--5215.

\bibitem[\protect\citeauthoryear{Ouali et~al.}{2023}]{ouali2023lfa}
Ouali Y, Bulat A, Matinez B, Tzimiropoulos G.
\newblock Black box few-shot adaptation for vision-language models.
\newblock In: ICCV; 2023. p. 15534--15546.

\bibitem[\protect\citeauthoryear{Lee et~al.}{2023}]{lee2023rpo}
Lee D, Song S, Suh J, Choi J, Lee S, Kim HJ.
\newblock Read-only prompt optimization for vision-language few-shot learning.
\newblock In: ICCV; 2023. p. 1401--1411.

\end{thebibliography}

\begin{appendices}
\section{Additional
Results}\label{secA1}

\begin{table*}[htbp]
\setlength{\abovecaptionskip}{0.cm}  
\setlength{\belowcaptionskip}{0.1cm} 
  \centering 
  \tabcolsep 0.008in
  \footnotesize
  \caption{Cross-dataset generalization performance of state-of-the-art PT methods w/ or w/o our CaPT on 11 datasets.}
    \begin{tabular}{l|c|ccccccccccc}
       \hline
     \multicolumn{1}{l|}{\multirow{2}{*}{\makecell[c]{{Method}}}}  & {{Source}}  & \multicolumn{10}{c}{{Target}}   \\
    \cline{2-13} 
    & \multicolumn{1}{c|}{ImgNet} & Calt.101 & {Oxford.} & {Stanford.} & {Flowers.} & {Food101}& {FGVCAir.}& {SUN397}& {DTD}& {EuroSAT}& {UCF101}& {{\textbf{Avg}}} \\
     \hline
     {CoOp} \cite{zhou2022coop}, \textit{IJCV22}  &\textbf{71.47}&\textbf{94.07}&90.07&64.97&69.60&86.00&21.07&66.37&43.50&\textbf{48.70}&68.13&65.25 \\
     \textbf{+CaPT}(\textit{Ours})  &71.20&93.97&\textbf{90.70}&\textbf{66.00}&\textbf{71.77}&\textbf{86.07}&\textbf{22.67}&\textbf{67.20}&\textbf{44.80}&42.73&\textbf{69.33}&\textbf{65.52} \\
    \hline
    {KgCoOP} \cite{yao2023kgcoop}, \textit{CVPR23}  &{71.11}&93.64&90.02&64.13&69.44&85.99&21.14&66.81&45.45&42.68&67.71&64.70 \\
     \textbf{+CaPT}(\textit{Ours})  &{\textbf{71.54}}&\textbf{94.02}&\textbf{90.91}&\textbf{65.18}&\textbf{71.13}&\textbf{87.12}&\textbf{22.05}&\textbf{66.92}&\textbf{46.27}&\textbf{42.80}&\textbf{69.11}&\textbf{65.55} \\
     \hline
    {ProGrad} \cite{zhu2022prograd}, \textit{ICCV23}  &69.37&93.87&89.77&\textbf{65.73}&70.30&\textbf{85.97}&\textbf{23.63}&65.77&45.27&\textbf{43.97}&67.00&65.13 \\
     \textbf{+CaPT}(\textit{Ours})  &\textbf{69.97}&\textbf{93.97}&\textbf{89.87}&65.57&\textbf{70.90}&85.83&23.50&\textbf{65.90}&\textbf{45.70}&43.43&\textbf{67.40}&\textbf{65.21} \\
     \hline
    {DAPT} \cite{zhang2023dept}, \textit{ICCV23}  &\textbf{71.77}&93.63&90.37&65.03&69.30&85.67&21.43&65.27&41.30&42.13&68.13&64.23 \\
     \textbf{+CaPT}(\textit{Ours})  &71.00&\textbf{94.07}&\textbf{90.57}&\textbf{65.77}&\textbf{70.77}&\textbf{86.13}&\textbf{23.07}&\textbf{66.80}&\textbf{44.43}&\textbf{43.73}&\textbf{68.30}&\textbf{65.36} \\
     \hline
    {DePT} \cite{zhang2023dept}, \textit{CVPR24}(\textit{Ours})  &72.80&93.63&\textbf{90.27}&\textbf{65.83}&70.63&\textbf{86.43}&22.30&66.70&45.50&43.20&69.13&65.36 \\
     \textbf{+CaPT}(\textit{Ours})  &\textbf{73.37}&\textbf{94.37}&90.07&65.57&\textbf{71.03}&86.37&\textbf{23.43}&\textbf{67.10}&\textbf{46.07}&\textbf{43.77}&\textbf{69.23}&\textbf{65.70} \\
      \hline
    \end{tabular}
  \label{table—CD}
\end{table*}
\begin{table}[tbp]
\setlength{\abovecaptionskip}{0.cm}  
\setlength{\belowcaptionskip}{0.1cm} 
  \centering 
  \tabcolsep 0.0034in
  \footnotesize
 \caption{Domain generalization performance of state-of-the-art PT methods w/ or w/o CaPT on 11 datasets.}
    \begin{tabular}{l|c|cccc}
       \hline
     \multicolumn{1}{l|}{\multirow{2}{*}{\makecell[c]{{Method}}}}  & {{Source}}  & \multicolumn{4}{c}{{Target}}   \\
    \cline{2-6} 
    &  \multicolumn{1}{c|}{ImgNet} & -V2 & -Sk. & -A & -R \\
     \hline
     {CoOp}  \cite{zhou2022coop}, \textit{IJCV22}  &\textbf{71.47}&\textbf{64.53}&48.77&50.50&76.37\\
     \textbf{+CaPT}(\textit{Ours})  &71.20&64.23&\textbf{49.00}&\textbf{50.67}&\textbf{76.57}\\
    \hline
    {KgCoOp} \cite{yao2023kgcoop}, \textit{CVPR23}   &{71.11}&{63.89}&48.80&49.89&{76.13}\\
     \textbf{+CaPT}(\textit{Ours})  &\textbf{71.54}&\textbf{64.07}&\textbf{49.33}&\textbf{50.11}&\textbf{76.55}\\
    \hline
    {ProGrad} \cite{zhu2022prograd}, \textit{ICCV23}   &69.37&62.60&47.57&\textbf{49.37}&75.70\\
     \textbf{+CaPT}(\textit{Ours})  &\textbf{69.97}&\textbf{63.07}&\textbf{47.73}&\textbf{49.37}&\textbf{75.87}\\
    \hline
    {DAPT} \cite{zhang2023dept}, \textit{ICCV23}   &\textbf{71.77}&\textbf{64.60}&47.83&49.60&75.53\\
     \textbf{+CaPT}(\textit{Ours})  &71.00&64.20&\textbf{48.37}&\textbf{50.50}&\textbf{76.43}\\
    \hline
    {DePT} \cite{zhang2023dept}, \textit{CVPR24}(\textit{Ours}) &72.80&64.70&47.57&48.37&75.10\\
     \textbf{+CaPT}(\textit{Ours})  &\textbf{73.37}&\textbf{65.10}&\textbf{48.20}&\textbf{49.10}&\textbf{75.77}\\
      \hline
    \end{tabular}
  \label{table—DG}
\end{table}

\keypoint{Cross-Dataset Generalization.}
The cross-dataset generalization setting assesses whether models trained on the source dataset (i.e., the base task)  can generalize to unseen target datasets (i.e., new tasks)—indicating there exists \textit{distribution shifts} between the base task and new tasks. 
This experiment follows the baseline approaches to use ImageNet as the source dataset and the other 10 datasets as target datasets. The achieved cross-dataset generalization results of the five baseline methods with or without CaPT across the 11 datasets are represented in Table \ref{table—CD}. 
From the average results, we observe that our CaPT consistently enhances the performance of the five baselines on the 10 target datasets, without compromising the performance of the tuned models on the source dataset in most cases. 
This demonstrates the flexibility and effectiveness of CaPT for improving the robustness of existing PT schemes to distribution shifts. 
Besides, by applying CaPT to our recently proposed PT method DePT, we further boost the state-of-the-art {cross-dataset generalization} results over the 11 datasets.

\keypoint{Domain Generalization.}
The domain generalization setting assesses whether models trained on the source domain (i.e., the base task) can generalize to unseen target domains (i.e., new tasks) —indicating there exists \textit{domain shift}s between the base task and new tasks. 
In line with the baseline approaches, we consider the ImageNet dataset as the source domain and the other four ImageNet variants as target domains.
The achieved domain generalization results of the five baseline methods with or without CaPT on the ImageNet dataset are represented in Table \ref{table—DG}. 
As seen, CaPT still maintains the advantages as in previous experiments. 
CaPT enhances the performance of most baseline methods on the source domain and target domains, proving the flexibility and effectiveness of CaPT for improving the robustness of existing PT approaches to domain shifts. 
Besides, we notice that the performance improvement established by CaPT in Table \ref{table—DG} are not as significant as those in Tables \ref{table—B2N} and \ref{table—CD}. 
Possible reasons are twofold.
On the one hand, domain generalization is a more challenging task compared to base-to-new generalization and cross-dataset generalization, as validated in \cite{zhou2022conditional,yao2023kgcoop}.
On the other hand, capturing both task-specific and domain-agnostic knowledge during PT is challenging without access to the data from target domains.

\end{appendices}


\end{document}